\documentclass[10pt, conference, compsocconf]{IEEEtran}
\ifCLASSINFOpdf
\else
\fi
\newif\ifdraft
\drafttrue

\newcommand{\myR}{\mathbb{R}}
\newcommand{\parag}[1]{{\vspace{2pt}\noindent\bf{#1}}}

\newcommand{\vd}{\mathbf{d}}

\newcommand{\vk}{\mathbf{k}}

\newcommand{\vp}{\mathbf{p}}

\newcommand{\vw}{\mathbf{w}}
\newcommand{\vx}{\mathbf{x}}

\newcommand{\vz}{\mathbf{z}}

\newcommand{\mD}{\mathbf{D}}

\newcommand{\mF}{\mathbf{F}}

\newcommand{\mH}{\mathbf{H}}
\newcommand{\mI}{\mathbf{I}}

\newcommand{\mM}{\mathbf{M}}

\newcommand{\mS}{\mathbf{S}}

\newcommand{\cD}{\mathcal D}

\newcommand{\cG}{\mathcal G}

\newcommand{\cK}{\mathcal K}
\newcommand{\cL}{\mathcal L}

\newcommand{\cN}{\mathcal N}

\newcommand{\cS}{\mathcal S}

\usepackage{amsmath}%
\usepackage{wasysym}
\usepackage{graphicx} 
\usepackage{multirow}%

\usepackage{amssymb,amsfonts}%
\usepackage{amsthm}%
\usepackage{mathrsfs}%
\usepackage[figuresright]{rotating}%
\usepackage{xcolor}%
\usepackage{textcomp}%
\usepackage{manyfoot}%
\usepackage{booktabs}%
\usepackage{algorithm}%
\usepackage{algorithmicx}%
\usepackage{algpseudocode}%
\usepackage{program}%
\usepackage{listings}%
\usepackage{hyperref}

\begin{document}
%
% paper title
% can use linebreaks \\ within to get better formatting as desired
\title{LatentKeypointGAN: Controlling Images via Latent Keypoints}

% author names and affiliations
% use a multiple column layout for up to two different
% affiliations

\author{\IEEEauthorblockN{Xingzhe He, Bastian Wandt, Helge Rhodin}
\IEEEauthorblockA{Computer Science Department\\
University of British Columbia\\
Vancouver, Canada\\
{xingzhe, wandt, rhodin}@cs.ubc.ca}
}

% conference papers do not typically use \thanks and this command
% is locked out in conference mode. If really needed, such as for
% the acknowledgment of grants, issue a \IEEEoverridecommandlockouts
% after \documentclass

% for over three affiliations, or if they all won't fit within the width
% of the page, use this alternative format:
% 
%\author{\IEEEauthorblockN{Michael Shell\IEEEauthorrefmark{1},
%Homer Simpson\IEEEauthorrefmark{2},
%James Kirk\IEEEauthorrefmark{3}, 
%Montgomery Scott\IEEEauthorrefmark{3} and
%Eldon Tyrell\IEEEauthorrefmark{4}}
%\IEEEauthorblockA{\IEEEauthorrefmark{1}School of Electrical and Computer Engineering\\
%Georgia Institute of Technology,
%Atlanta, Georgia 30332--0250\\ Email: see http://www.michaelshell.org/contact.html}
%\IEEEauthorblockA{\IEEEauthorrefmark{2}Twentieth Century Fox, Springfield, USA\\
%Email: homer@thesimpsons.com}
%\IEEEauthorblockA{\IEEEauthorrefmark{3}Starfleet Academy, San Francisco, California 96678-2391\\
%Telephone: (800) 555--1212, Fax: (888) 555--1212}
%\IEEEauthorblockA{\IEEEauthorrefmark{4}Tyrell Inc., 123 Replicant Street, Los Angeles, California 90210--4321}}

% use for special paper notices
%\IEEEspecialpapernotice{(Invited Paper)}

% make the title area
\maketitle

% For peer review papers, you can put extra information on the cover
% page as needed:
% \ifCLASSOPTIONpeerreview
% \begin{center} \bfseries EDICS Category: 3-BBND \end{center}
% \fi
%
% For peerreview papers, this IEEEtran command inserts a page break and
% creates the second title. It will be ignored for other modes.
\IEEEpeerreviewmaketitle

\begin{abstract}
Generative adversarial networks (GANs) can now generate photo-realistic images. However, how to best control the image content remains an open challenge. We introduce LatentKeypointGAN, a two-stage GAN internally conditioned on a set of keypoints and associated appearance embeddings providing control of the position and style of the generated objects and their respective parts. A major difficulty that we address is disentangling the image into spatial and appearance factors with little domain knowledge and supervision signals. We demonstrate in a user study and quantitative experiments that LatentKeypointGAN provides an interpretable latent space that can be used to re-arrange the generated images by re-positioning, adding, removing, and exchanging keypoint embeddings, such as generating portraits by combining the eyes, and mouth from different images. Notably, our method does not require labels as it is self-supervised and thereby applies to diverse application domains, such as editing portraits, indoor rooms, and full-body human poses. In addition, the explicit generation of keypoints and matching images enables a new, GAN-based method for unsupervised keypoint detection.
\end{abstract}

\begin{figure*}
% \vspace{-5pt}
% \setlength{\abovecaptionskip}{-2pt}
\centering
   \includegraphics[width=0.85\linewidth]{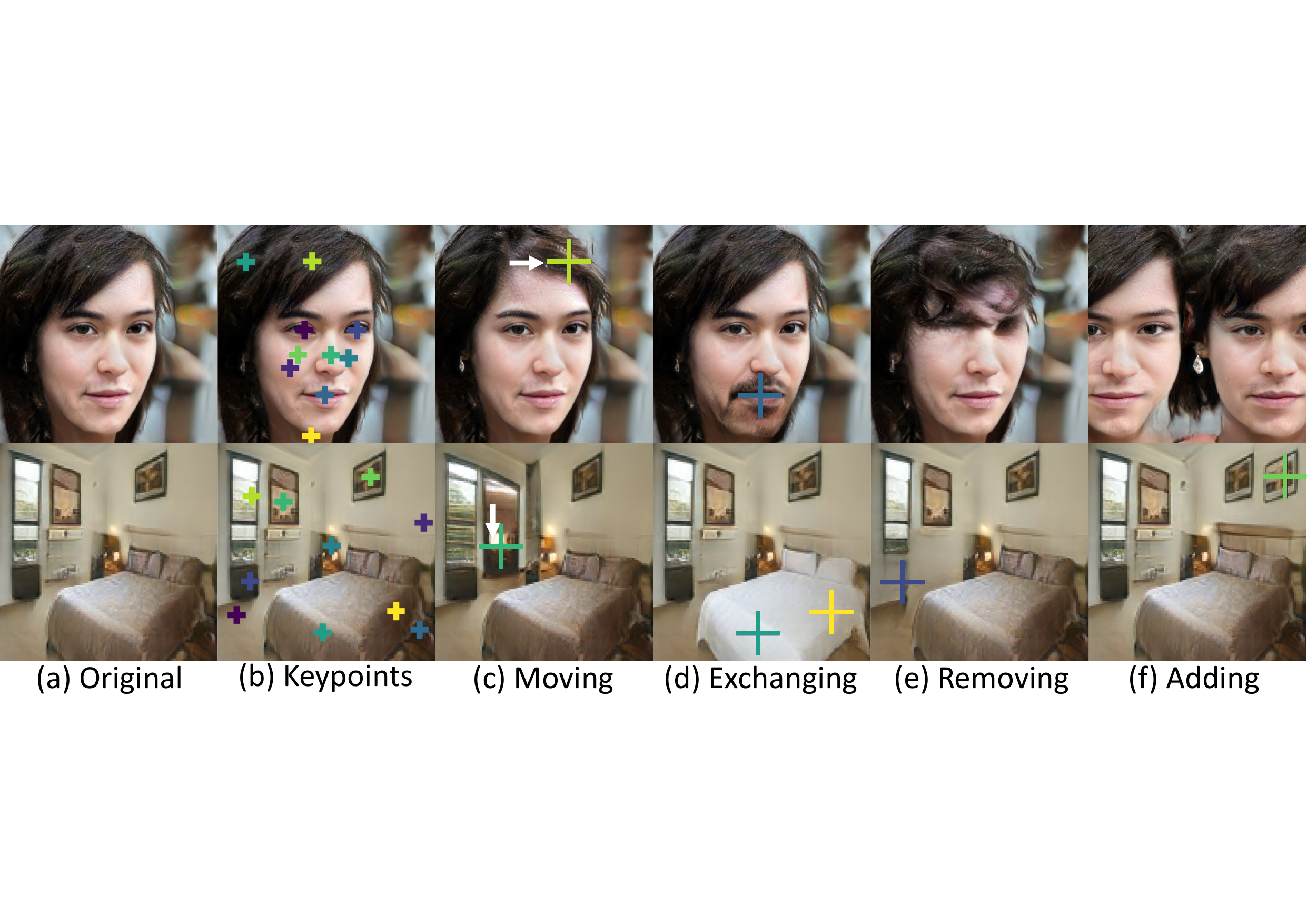}
\caption{GANs can generate phot-realistic images (a) but lack local editing capability. \textbf{LatentKeypointGAN} generates images with associated keypoints (a-b), which enables local editing by moving keypoints (c), exchanging appearance (d), removing individual parts (e), and adding one or more parts (f). Our improvements are on the unsupervised learning of an interpretable latent space that disentangles pose and appearance, which makes it easy to use and applicable to diverse domains, including portraits (top row), indoor rooms (bottom row), and persons (see results section).}%
\label{fig:teaser}
% \vspace{-5pt}
\end{figure*}
\section{Introduction} \label{sec:intro}

\vspace{0.01cm}
It is a long-standing goal to build generative models that produce photo-realistic images driven by intuitive user input. 
While photo-realism is already reached for well-constrained domains, such as portraits, it remains challenging to make this image generation process interpretable and editable. 
Desired is a latent space that disentangles an image into parts and their appearances, which would allow a user to re-combine and re-imagine a generated portrait interactively. 

There are promising methods
%in the area of unsupervised keypoint detection
\cite{zhang2018unsupervised, jakab2018unsupervised, lorenz2019unsupervised}
that use spatial image transformations, such as the thin-plate spline (TPS) method, to create pairs of real and deformed images and impose an equivariance loss to discover keypoints and object appearance embeddings as the bottleneck of an autoencoder. 
Thereby, one can edit the image by moving, the keypoints or modifying the appearance embedding. 
Yet, generated images lack fine detail. On the other hand, GAN-based editing approaches attain impressive image quality but lack some editing capabilities. 
Methods such as StyleGAN~\cite{karras2019style, karras2020analyzing} and SPADE~\cite{park2019semantic} enable the mixing of appearance properties from different faces and to synthesize natural images by 'painting' source and target regions in the feature space. 
While powerful, their editing on feature maps makes it difficult to faithfully reposition image parts spatially as the influence region of the mask has to be drawn manually with pixel-level accuracy. Our goal is user-friendly control via automatically learned keypoints providing handles analogous to how character rigs are keyframed in classical animation, thereby overcoming manual drawing and applying to domains without semantic labels.

Inspired by the control of autoencoder-based techniques and by the improved image quality of GANs, we introduce keypoint locations and associated feature embeddings as latent variables in the generator network of a GAN. Thereby, the location and appearance of image parts is separated and can be controlled. Figure~\ref{fig:teaser} shows how \emph{LatentKeypointGAN} enables editing the output image by changing the keypoint position, adding or removing points, and exchanging associated appearance embeddings locally while maintaining a high image quality that approaches that of existing GANs. 

We target an unsupervised setting in that the position, extend, and appearance of \emph{parts} (image regions that share appearance and belong together) is learned from unlabelled example images. It eases the application to new domains, where large image collections are available but exact segmentation masks or part labels are missing.
By using the GAN objective in favor of image quality, we cannot rely on the equivariance constraints that are established for autoencoders setups. Instead, we introduce new auxiliary objective functions and control the flow of information in the network to re-instantiate equivariance properties and to disentangle pose and appearance. 

LatentKeypointGAN is designed as a two-stage GAN architecture that is trained end-to-end. In the first step, a generator network turns the input values sampled from a normal distribution into 2D keypoint locations and their associated encoding. We ensure with suitable neural network pathways that some of the encodings are correlated while others remain independent. These generated keypoints are then mapped to spatial heatmaps of increasing resolution. The heatmaps define the position of the keypoints and their support sets the influence range of their respective encodings. In the second step, a SPADE-like \cite{park2019semantic} image generator turns these spatial encodings into a complete and realistic image. 
Although entirely unsupervised, the learned keypoints meaningfully align with the image landmarks, such as a keypoint linked to the nose when generating images of faces, enabling the desired editing. As a byproduct, we can learn a separate keypoint detector on generated image-keypoint pairs for unsupervised keypoint detection, which we utilize to quantify localization accuracy.

We summarize our contributions below: 
\begin{enumerate}
    \item A GAN-based framework for handle-based image manipulation requires less user input than existing techniques and that succeeds on more diverse domains;
    % \item Design of a keypoint generator that models dependent and independent factors explicitly;
    \item A new GAN-based methodology for keypoint detection that contests established autoencoder methods;
    % \item Empirical study comparing different editing methods in terms of perceptual quality;
    \item A new metric to compare part disentanglement across existing models.
\end{enumerate}

The supplemental materials are \textbf{\href{https://drive.google.com/drive/folders/1vzzMWQAEJpxJXD3M7_7H9B7aptIwSvb6?usp=sharing/}{here}}.
\section{Related Work} \label{sec:related_work}

% In the following, we discuss variants of generative models that learn to synthesize images from a collection of examples, focusing on methods providing control on the image content. Table~\ref{tab:features} summarizes the editing features of the most-related methods.
In the following, we focus on methods providing local editing with unconditional GANs. Table~\ref{tab:features} summarizes the editing features of the most-related methods.
\newcommand*\rot{\rotatebox{90}}

\begin{table}[t]
\centering
\resizebox{\linewidth}{!}
{
\begin{tabular}{ l r r r r r r r r r r}
%\toprule
Feature
& {\bf \rot{Zhang et al.~\cite{zhang2018unsupervised}}} 		
& {\bf \rot{Lorenz et al.~\cite{lorenz2019unsupervised}}} 		
& {\bf \rot{Karras et al.~\cite{karras2020analyzing}}} 		
& {\bf \rot{Collins et al.~\cite{collins2020editing}}} 	
& {\bf \rot{Alharbi et al.~\cite{alharbi2020disentangled}}} 		
& {\bf \rot{Kim et al.~\cite{kim2021exploiting}}} 		
& {\bf \rot{Wang et al.~\cite{wang2018high}}} 		
& {\bf \rot{Park et al.~\cite{park2019semantic}}} 		
& {\bf \rot{Zhu et al.~\cite{zhu2020sean}}}
& {\bf \rot{Ours} }		\\
\midrule
% These are control issues
Appearance transfer	(global) & \CIRCLE & \CIRCLE	& \CIRCLE 	& \RIGHTcircle	& \CIRCLE		& \CIRCLE 		& \CIRCLE 		& \CIRCLE 		& \CIRCLE 	& \CIRCLE 		\\
Appearance transfer (local, part-based) & \CIRCLE & \CIRCLE  & \Circle 	& \RIGHTcircle 	& \CIRCLE 		& \CIRCLE 		& \CIRCLE 		& \CIRCLE 		& \CIRCLE 	& \CIRCLE 		\\
Removing and adding parts & \CIRCLE  & \CIRCLE  & \Circle 	& \Circle 	& \Circle 		& \Circle 		& \CIRCLE 		& \CIRCLE 		& \CIRCLE 	& \CIRCLE 		\\
Moving parts spatially 	 & \CIRCLE 	& \CIRCLE	& \Circle 	& \Circle 	& \Circle 		& \Circle 		& \Circle 		& \Circle 		& \Circle 	& \CIRCLE 		\\
Image quality w/o edits	& \Circle 	& \Circle & \CIRCLE &	\CIRCLE 	& \CIRCLE 		& \CIRCLE 		& \RIGHTcircle 		& \RIGHTcircle 		& \CIRCLE 	& \RIGHTcircle 		\\
Image quality after editing	& \Circle 	& \Circle  & \CIRCLE &	\CIRCLE 	& \CIRCLE 		& \CIRCLE 		& \RIGHTcircle 		& \RIGHTcircle 		& \RIGHTcircle 	& \RIGHTcircle 		\\ % modified the third-last value from 1/4 to 1/2 \pie{90} -> \RIGHTcircle
Training w/o part annotation (unsupervised)	& \CIRCLE &	\CIRCLE   & \CIRCLE &	\CIRCLE 	& \CIRCLE 		& \CIRCLE 		& \Circle 		& \Circle 		& \Circle 	& \CIRCLE 		\\
Inference w/o manual feature region 'painting'	& \CIRCLE &	\CIRCLE & \CIRCLE &	\CIRCLE 	& \Circle 		& \Circle 		& \CIRCLE 		& \CIRCLE 		& \CIRCLE 	& \CIRCLE 		\\
\bottomrule
\end{tabular}
}
\caption{$\CIRCLE$ / $\RIGHTcircle$ / $\Circle$ : full / partial / no support; 
Feature table comparison to state-of-the-art generative image editing methods. %
% For our task of intuitive editing, our method is a better fit than existing methods, though sacrificing some image quality over GANs with less control---an inevitable trade-off.
}
\label{tab:features}
\end{table}
GANs ~\cite{goodfellow2014generative} are trained to generate images from a distribution that resembles the training distributions. Recent approaches attain photo-realism for images \cite{karras2018progressive, karras2019style, karras2020analyzing} and gain some control by 
modyfying latent features.
Instead of exchanging entire feature maps, local modifications are possible but require care to maintain consistent and artifact-free synthesis. 
\cite{collins2020editing} propose to cluster regions in the feature maps, but this requires to select suitable clusters. 
\cite{kwon2021diagonal} manipulate attention maps, but they do not demonstrate local appearance editing.
\cite{alharbi2020disentangled} and~\cite{kim2021exploiting} inject structured noise to feature maps at inference, but they requires hand-pick regions to edit for every instance. 
\cite{niemeyer2021giraffe} learn in an unsupervised manner to modify objects but not on modeling parts.
\cite{he2022ganseg} achieves better location accuracy for articulated objects but is not suited for editing.
\cite{oldfield2023panda} decompose the feature map and achieve great local editing but it fails to capture the light and shadow.
Differently, our model provides an \textit{explicitly} high-level control to change the pose by modifying keypoint locations that are consistent across images. While enabling very useful forms of editing, all of the methods above require some form of a pixel-level selection of regions at test time or manual cluster selection.
As a result, they provide limited spatial editing control. Neither of them demonstrates moving parts relative to each other or adding and removing parts. In sum, our keypoint-based editing is both more intuitive and enables additional editing capabilities. \looseness=-1

% \section{Algorithm Overview} \label{sec:method}

\begin{figure*}
\begin{center}
\includegraphics[width=0.95\linewidth]{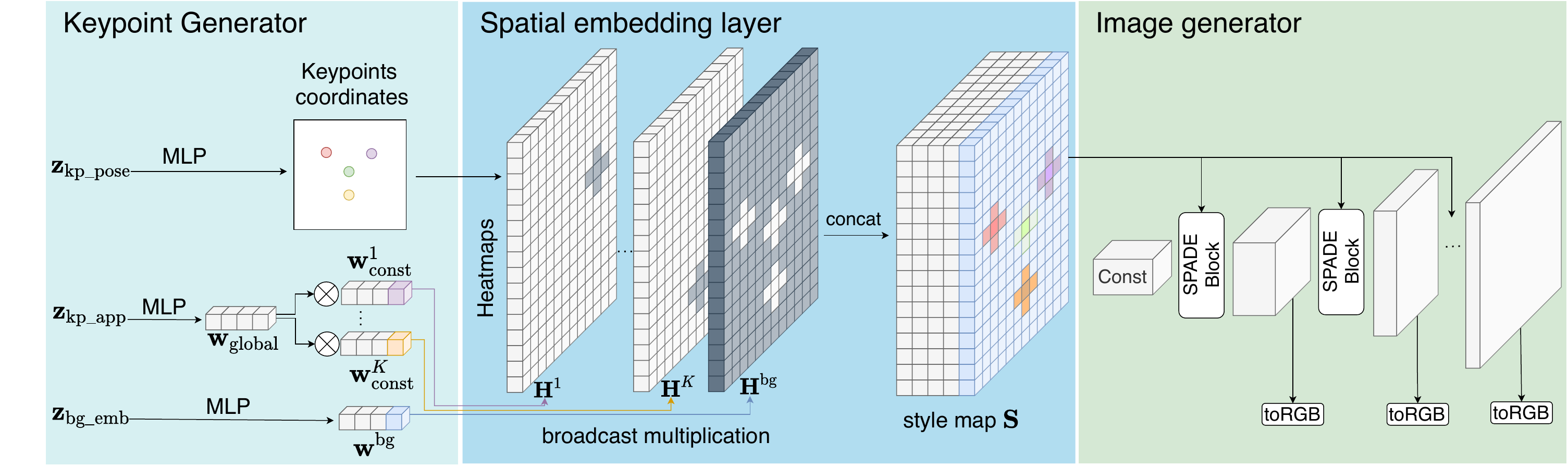}
% \vspace{-10pt}
\end{center}
   \caption{\textbf{Overview.} Starting from noise $\vz$, LatentKeypointGAN generates keypoint coordinates, $\vk$ and their embeddings $\vw$. Cruicial is how they are turned into feature maps that are localized around the keypoints, forming conditional maps for the image generation via SPADE block at different resolutions. At inference time, the position and embedding of keypoints can be edited by the user to control the position and appearance of parts.
   %SPADE-LeakyReLU-Conv. 
%   High-resolution images are generated with progressive training, conditioned on increasingly large feature maps.
   %The images of different resolutions for progressive training are generated by feature maps of different resolutions using $1\times 1$ convolution. The final images for testing are generated from the feature map with largest resolution. 
   }
\label{fig:archi}
\end{figure*}

\section{Method}

Given a set of example images, our aim is a GAN that starting from random noise generates new images of the same type, e.g., indoor or portrait, and provides control over meaningful image parts.
% by parametrizing the image in terms of keypoint locations $p$ and their embeddings $w$. 
We operate in the unsupervised setting, these parameters are latent variables that are inferred from the example images without requiring labels such as segmentation masks. 
Our proposed architecture operates in two stages. Figure \ref{fig:archi} shows the entire architecture.
First, the keypoint generator, $\cK$, defines the spatial arrangement of parts and their embedding explicitly as a point cloud $\{p^j\}_{j=1}^K$ and associated point-wise feature embeddings $\{w^j\}_{j=1}^K$. Its responsibility is to learn the spatial arrangement and part-specific appearance.
In the next stage, a spatial embedding layer, $\cS$, turns these sparse estimates into dense feature maps that are amendable for processing by the convolutional image generator $\cG$. The generator up-scales features into an image. 
At inference time, the latent keypoints allow one to author the keypoint location and appearance interactively.

\subsection{Keypoint Generator}
\label{sec: keypoint generator}

The keypoint generator $\cK$ learns the embeddings and spatial arrangement of image parts, such as eyes, nose, and mouth for describing a portrait image. 
It takes three equally shaped Gaussian noise vectors as input $\vz_\text{kp\_pose},\vz_\text{kp\_app}, \vz_\text{bg\_emb} \sim \mathcal N(\mathbf 0^{D_\text{noise}}, \mI^{D_\text{noise}\times D_\text{noise}})$, where $D_\text{noise}$ is the noise dimension. Each vector is passed through a three-layer MLP to respectively 
generate the $K$ keypoint coordinates $\vk^j\in[-1,1]^2, j=1,...,K$, a global style vector $\vw_\text{global}\in\myR^{D_\text{embed}}$, and a background embedding $\vw_\text{bg}$.
%The MLP can learn dependencies, e.g., correlation between keypoint locations.
Here $K$ is a pre-defined hyperparameter.
Crucial for the desired pose-appearance-disentanglement and part-disentanglement is that the noise for all three factors is independent and how the keypoint embedding $\vw^j$ is combined with a global embedding. We found that learning a constant factor during training that is multiplied with the varying input noise works best. Formally we write,
\begin{equation}
    \vw^j=\vw_\text{global} \otimes \vw^j_\text{const},
    \label{eq:noise combination}
\end{equation}
with $\otimes$ the elementwise product. The constant embedding $\vw^j_\text{const}\in\myR^{D_\text{embed}}$ is designed to
encode the keypoint semantics, e.g., left or right eye. They are updated during the training but fixed during inference. The
global style vector $\vw_\text{global}\in\myR^{D_\text{embed}}$ can be regarded as 
learning the correlation of parts while the global noise ensures that a different appearance is drawn for every generated image. 
% We show in the evaluation that this factorization into dependent and independent factors is nontrivial and improves on straightforward variants that directly use an MLP.

\subsection{Spatial Embedding Layer}
\label{sec: embedding layer}
With keypoint coordinates and embeddings generated, we now turn these point-wise estimates into an image. 
To this end, we generate Style maps $\mS^j\in\myR^{D_\text{embed}\times H\times W}$ for each keypoint $j$,
% These are low-resolution prototypes with the embedded features describing the part appearance and the position of the features localized to the corresponding keypoint position. Thereby the subsequent convolutional generator combines nearby part features into a crisp image, maintaining the desired locality and translation invariance. Style maps are generated 
by multiplying the keypoint embedding $\vw^j\in\myR^{D_\text{embed}}$ with a Gaussian heatmap $\mH^j\in\myR^{H\times W}$ that defines the local support. For a pixel $\vp$
\begin{equation}
    \begin{aligned}
       \mS^j(\vp)&=\mH^j(\vp)\vw^j, \\
       \text{ where }  \mH^j(\vp)&=\exp\left(-\|\vp-\vk^j\|_2^2/\tau \right)
     \end{aligned}
\label{eq:heatmap}
\end{equation}
has Gaussian shape, is centered at the keypoint location $\vk^j$, and $\tau$ controls the influence range. We also define a heatmap $\mH^\text{bg}$ for the background as the negative of all keypoint maps, $\mH^\text{bg}(\vp)=1-\max_j\{\mH^j(\vp)\}_{j=1}^K$, thereby ensuring the desired background separation. The background heatmap is multiplied with the independent noise vector $\vw^\text{bg}$ generated from $\vz_\text{bg\_emb}$ instead of keypoint embedding, but treated equally otherwise.
Then we concatenate all $K+1$ style maps 
%$\{\mS^j\}_{j=1}^K$ and the background style map $\mS^\text{bg}$ in the channel dimension 
to $\mS\in\myR^{ D_\text{embed}\times H\times W \times K+1}$. 
% How these are used as conditional variables at different levels of the image generator is explained in the next section.

\subsection{Image Generator} \label{sec:image_generator}

Our image generator $\cG$ follows the progressively growing architecture of StyleGAN \cite{karras2019style} and combines it with the idea of spatial normalization from SPADE \cite{park2019semantic}, which was designed to generate images conditioned on segmentation masks. 
Our generator starts from a learned $4\times4\times512$ constant tensor and keeps applying convolutions and upsampling to obtain feature maps of increasing resolution. 
% The generator increases resolution layer-by-layer with multiple adaptive normalization layers, requiring differently-sized feature maps.
We apply Equation~\ref{eq:heatmap} multiple times to generate style maps of different resolution.
% that replace the manually annotated segmentation masks in SPADE.

% To improve the image quality under editing operations, we replaced the BatchNorm \cite{ioffe2015batch} layers in \cite{karras2019style} with spatial adaptive normalization and removed the residual links \cite{he2016deep} because we found it detrimental in combination with progressive training.

\subsection{Loss Functions}

\parag{Adversarial losses.} We use the non-saturating loss \cite{goodfellow2014generative},
\begin{equation}
    \cL(\cG)_\text{GAN}=\mathbb E_{\vz\sim\cN}\log(\exp(-\cD(\cG(\vz)))+1)
\label{eq:gen_loss}
\end{equation}
for the generator, and logistic loss,
\begin{equation}
    \begin{aligned}
        \cL(\cD)_\text{GAN}=&\mathbb E_{\vz\sim\cN}\log(\exp(\cD(\cG(\vz)))+1)+\\
        &\mathbb E_{\vx\sim p_\text{data}}\log(\exp(-\cD(\vx))+1),
     \end{aligned}
\label{eq:dis_loss}
\end{equation}
for the discriminator, with gradient penalty \cite{mescheder2018training} applied only on real data,
\begin{equation}
    \cL(\cD)_\text{gp}=\mathbb E_{\vx\sim p_\text{data}}\nabla\cD(\vx).
\label{eq:gradient penalty}
\end{equation}
% Other losses such as hinge loss \cite{park2019semantic} failed.

\parag{Background loss.} To further disentangle the background and keypoints, and to stabilize their location, we introduce a background penalty,
\begin{equation}
\begin{aligned}
    \cL(\cG)_\text{bg}=\mathbb E_{\vz_1,\vz_2} [ &(1-\mH^\text{bg}_1) \otimes \cG(\vz_1)\\
    &-(1-\mH^\text{bg}_2) \otimes \cG(\vz_2)],
\end{aligned}
    \label{eq:bg_loss}
\end{equation}
where $\vz_1$ and $\vz_2$ share the same keypoint location and appearance input noise, and only differ at the background noise input. The $\mH_1$ and $\mH_2$ are the background heatmaps generated by $\vz_1$ and $\vz_2$. With this penalty, we expect the keypoint location and appearance do not change with the background.
The final loss for the discriminator and the generator are, respectively,
\begin{equation}
    \begin{aligned}
        \cL(\cD)&=\cL(\cD)_\text{GAN} + \lambda_\text{gp}\cL(\cD)_\text{gp},\\
        \cL(\cG)&=\cL(\cG)_\text{GAN} + \lambda_\text{bg}\cL(\cG)_\text{bg}.
     \end{aligned}
\end{equation}

\section{Results and Validation}   \label{experiments}

We compare our results on the established benchmarks to the most related methods in a variety of ways, showing that we introduce new editing capabilities while maintaining high image quality.

\subsection{Baselines}
We compare against the four most related ones providing editing (Tab.~\ref{tab:fid} bottom half), validating that we extend their editing capabilities while maintaining a high image quality after editing.
\textbf{Image translation methods} condition on a segmentation mask label, also at test time. To enable fair comparisons, we train a self-supervised keypoint detector (see below) to condition on test images whenever comparing to them. We compare against three methods (Tab.~\ref{tab:fid} top half), demonstrating equal image editing quality and similar disentanglement while not using any labels.
\textbf{Autoencoders} have been tailored for keypoint localization accuracy ~\cite{zhang2018unsupervised} and do not report FID metrics. We therefore only compare their quality in the user study and focus the quantitative comparison on localization accuracy.

\subsection{LatentKeypointGAN variants}
Unless specified otherwise, we set the scale parameter $\tau=0.01$ and use 10 keypoints. To determine localization accuracy and consistency, we learn an independent \textbf{self-supervised keypoint detector}. A standard ResNet detector \cite{xiao2018simple} supervised on 200,000 image-keypoint pairs generated on-the-fly by LatentKeypointGAN. We also tuned our architecture \textbf{LatentKeypointGAN-tuned} for precise localization by enforcing stronger translation equivariance. Please see the supplemental for details.

\begin{figure}[t]
\begin{center}
  \includegraphics[width=0.99\linewidth]{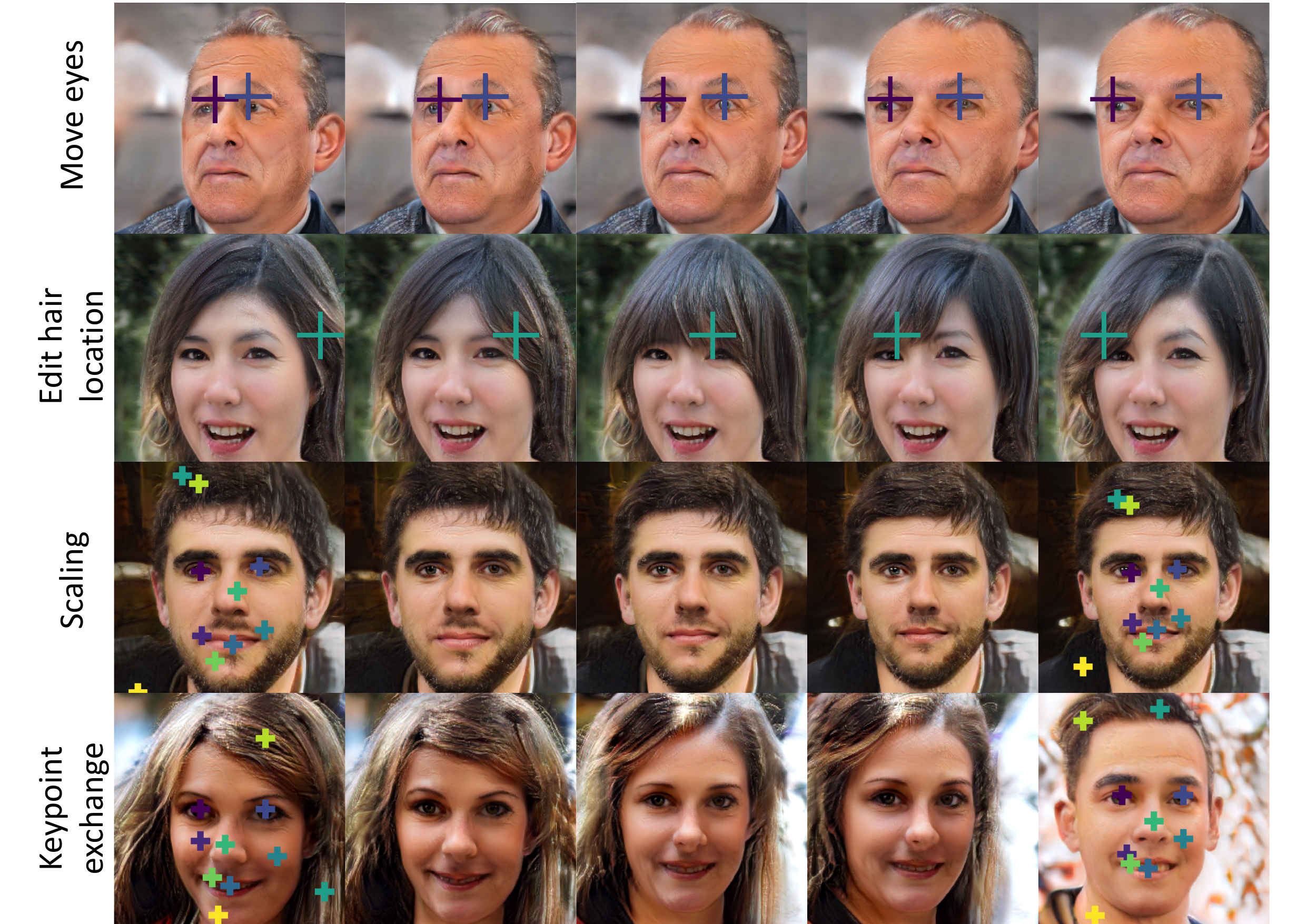}
  \vspace{-10pt}
\end{center}
  \caption{\textbf{Location and scale editing.} The first column is the source and the last the target. The images in-between are the result of the following operations.
  \textbf{First row:} pushing the eye keypoint distance from 0.8x to 1.2x. Note that the marked eye keypoints in this row are slightly shifted upward for better visualization.
  \textbf{Second row:} interpolating the hair keypoint to move the fringe from right to left. 
  \textbf{Third row:} scaling the keypoint location and, therefore, the face from 1.15x to 0.85x.
  \textbf{Fourth row:} interpolating all keypoint locations, to rotate the head to the target orientation. 
  }
\label{fig:reposition}
\end{figure}

\subsection{Metrics} We use the most widely used metrics, plus a new one on disentanglement and a user study on editing quality.

\parag{Image quality.} We use the Fr\'{e}chet inception distance (FID) \cite{heusel2017gans} 
% that compares the distribution of generated images with the distribution of real training images in the feature space of a pre-trained model. Scores are computed by using the Pytorch FID calculator \cite{Seitzer2020FID} 
caluclated on $N=50k$ generated and natural images.

\parag{Editing quality-global.} We use the $\text{FID}_\text{lerp}$ \cite{kim2021exploiting}, computed as the FID after editing. $N$ composite images are generated from random pairs of 500 generated images by linearly interpolating the latent variables (for ours embeddings and keypoint locations) of a pair. 

\parag{Editing quality-local.} 
% While $\text{FID}_\text{lerp}$ captures global editing, it does not apply to changing individual parts as incompatible parts would change the image distribution, distorting FID scores. Therefore, we compare local editing quality with a comparative user study.
Since $\text{FID}_\text{lerp}$ only captures global editing, we compare local editing quality with a user study.

\parag{Disentanglement.} \label{sec:cpd} Previous disentangling methods analyzed latent space trajectories~\cite{karras2019style}, which does not generalize across network architectures, or use the difference magnitude between an original and edited image over manually annotated segmentation masks~\cite{collins2020editing}, which is not applicable to mask-free methods, including our keypoint approach. To this end, we propose a new correlation part disentanglement (CPD) that generalizes better. We generate 2000 images and randomly pair them. For a model with $K$ parts, we then create $K$ variants by exchanging part embeddings one by one. We take the difference before and after editing and compute the spatial correlation of parts over all 1000 pairs. 
Figure~\ref{fig:CPD} shows how the off-diagonal entries of the resulting correlation matrix quantify how much two different parts overlap.
%\xz{this now may be questionable?} It is still an overlap, even though not necessary overlap of neighboring parts
Our CPD score captures this as one minus the average of the off-diagonal elements, which ranges from 1 (perfect disentanglement) to 0 (no control). For details, please refer to the supplemental document.

\subsection{Benchmarks} We use the official test splits from five different datasets. In the following, we detail the slightly varying protocol variants that are established for each.
For \textbf{portrait editing} we use: a resolution of $512\times 512$ on FFHQ~\cite{karras2019style} to compare to GAN approaches; $256\times256$ on CelebA-HQ~\cite{zhu2020sean} to match FID comparison with the imge translation methods, including~SEAN; and $128\times 128$ on CelebA~\cite{liu2015faceattributes} for the autoencoder methods. For editing experiments on FFHQ and detection experiments on CelebA, we augment by randomly cropping the training images to a size of 70-100\%.
For both \textbf{human pose} experiments on BBCPose \cite{charles2013domain} and the \textbf{indoor} domain on LSUN Bedroom \cite{yu2015lsun} the resolution is $128\times 128$.

\subsection{Interactive Editing}
\label{sec: qualitative results}

Our key advancement is the conditioning of a GAN on keypoint locations. Figure~\ref{fig:reposition} demonstrates how this enables moving of eyes and hair, and scaling of the entire face by scaling all keypoint positions. Surprisingly, due to the strong disentanglement of parts, the generator is not limited to the number of keypoints it is trained with. The teaser shows that removing and adding parts is possible by adding and removing an arbitrary number of peaks in the Gaussian heatmap and associated embedding masks. 
Figure~\ref{fig:functionality} compared editing capabilities to related GANs. None of these spatial operations have been demonstrated by existing GANs, as these focus on global and local editing via feature maps, but not on the repositioning and addition of parts as their parametrization on part position is not explicit. 
% \cite{zhu2020domain} comes closest to our editing results, they demonstrate global and local editing, and we used their public implementation to test adding and removing. However, they require segmentation masks at training time, and parts cannot be moved manually; the underlying masks would rather need to be repainted in a consistent manner, which is not applicable to interactive animation as we show for our method in the supplemental videos. Notably, the autoencoder methods offer the same editing as we do but at much lower image generation quality (see next section).

\begin{figure}[t]
\begin{center}
\includegraphics[width=0.99\linewidth]{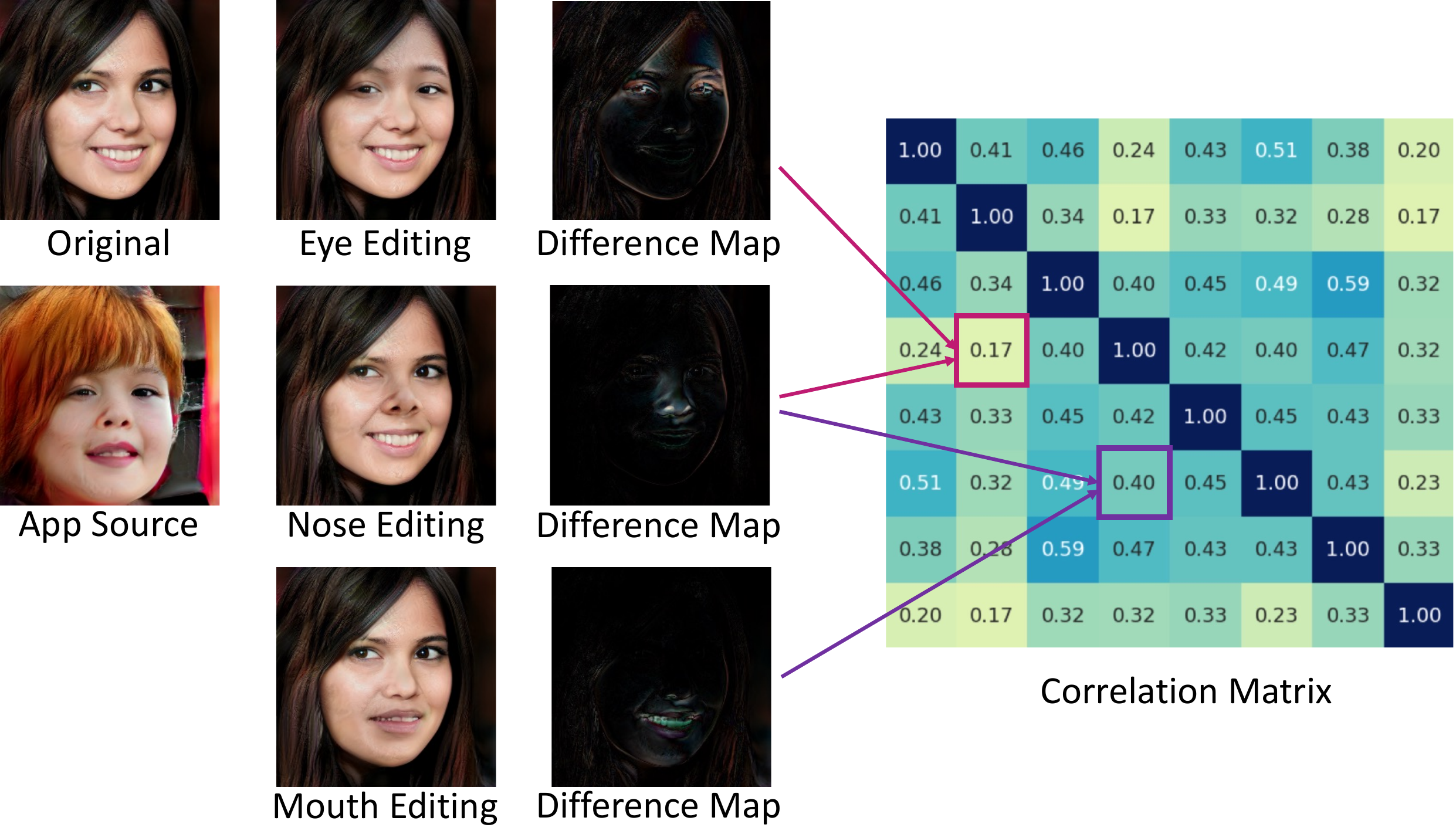}
\vspace{-10pt}
\end{center}
  \caption{\textbf{Our Correlation-based part Disentanglement (CPD) metric} is computed as the sum over part correlations. From a pair of images (left) the appearance of parts is exchanged one by one, here for eyes, nose, and mouth (center). The pairwise correlation of the resulting differences maps (right) forms the entries of the correlation matrix.}
\label{fig:CPD}
\end{figure}

\newcommand{\centertable}[1]{\begin{tabular}{@{}c@{}} \vspace{-2cm}\\#1\\$\phantom{space}$ \end{tabular}}
\newcommand{\cna}{\centertable{n/a}}

\begin{figure}[t]
\Huge
  \resizebox{0.98\linewidth}{!}{%
\begin{tabular}{cccccc@{}c@{}}
\centertable{\rot{\cite{kwon2021diagonal}}}&
\includegraphics[width=0.35\linewidth]{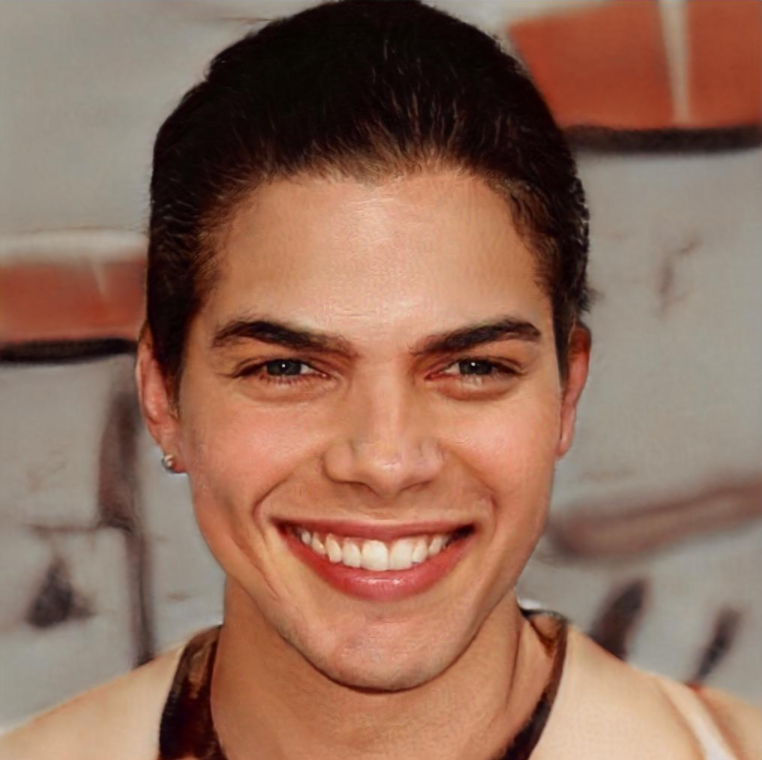}&
\includegraphics[width=0.35\linewidth]{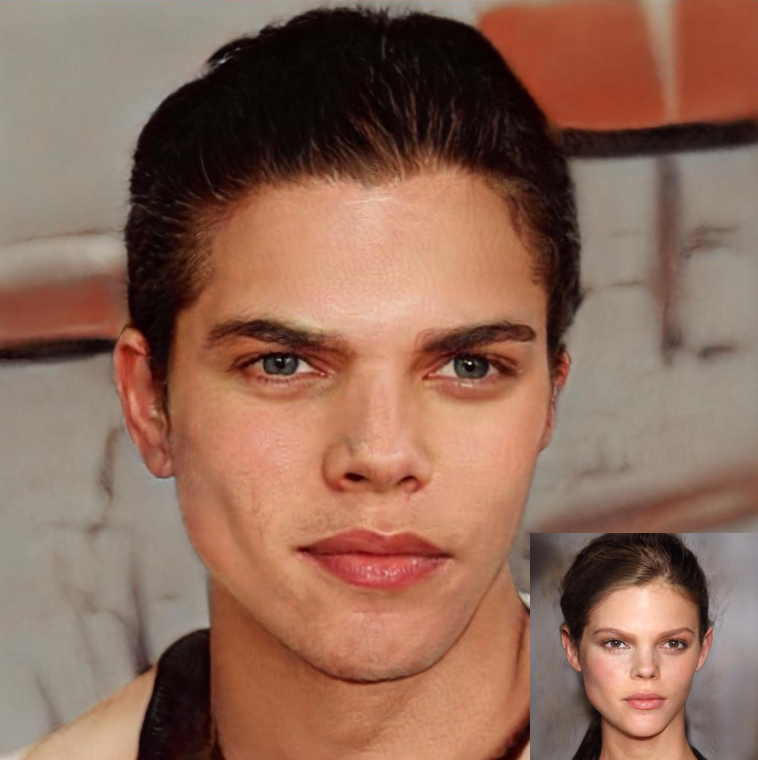}& \cna & \cna & \cna & \cna \\
\centertable{\rot{\cite{kim2021exploiting}}}&
\includegraphics[width=0.35\linewidth]{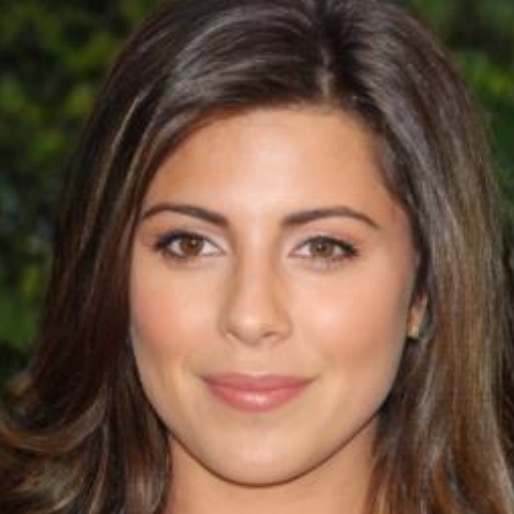}&
\includegraphics[width=0.35\linewidth]{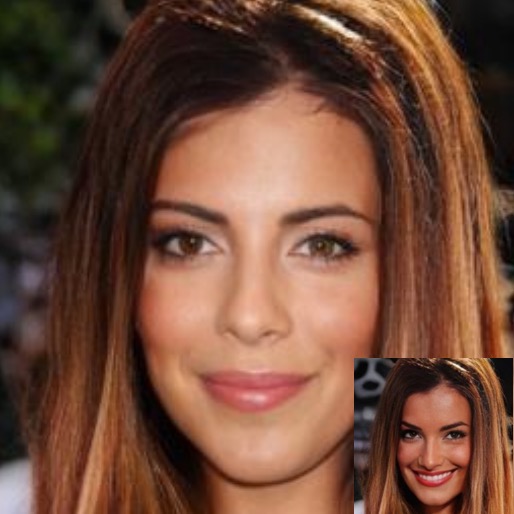}&
\includegraphics[width=0.35\linewidth]{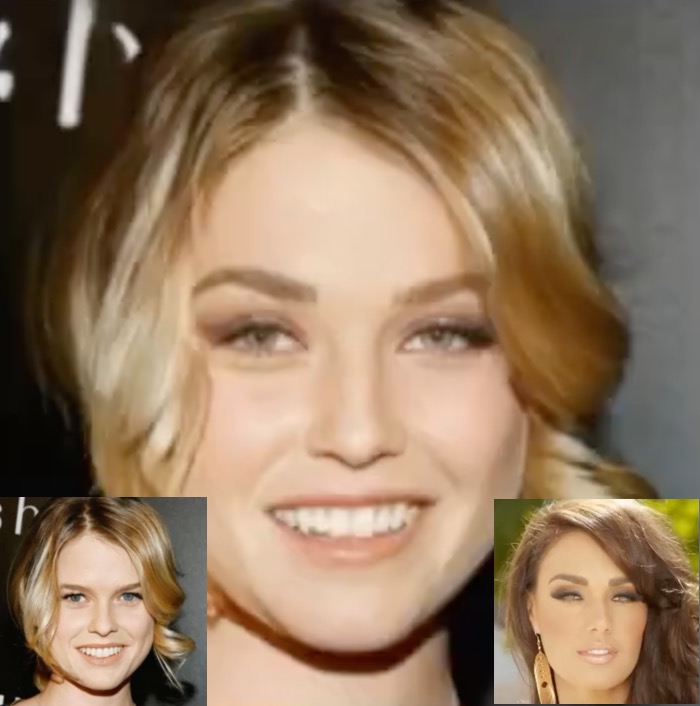}& \cna & \cna & \cna \\
\centertable{\rot{\cite{collins2020editing}}}&
\includegraphics[width=0.35\linewidth]{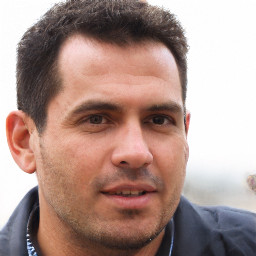}&
\includegraphics[width=0.35\linewidth]{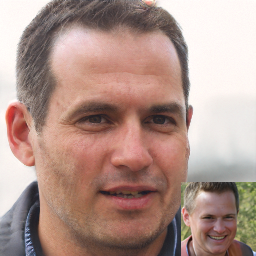}&
\includegraphics[width=0.35\linewidth]{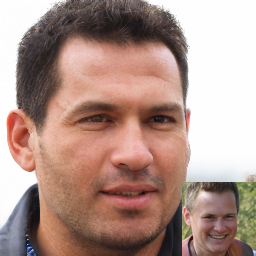}& \cna & \cna & \cna\\
\centertable{\rot{\cite{zhu2020sean}}}&
\includegraphics[width=0.35\linewidth]{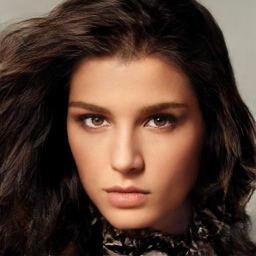}&
\includegraphics[width=0.35\linewidth]{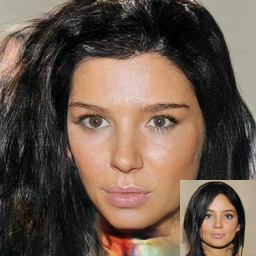}&
\includegraphics[width=0.35\linewidth]{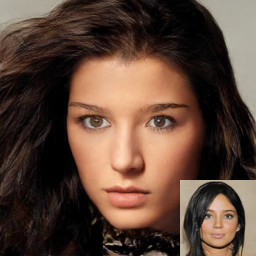}&
\includegraphics[width=0.35\linewidth]{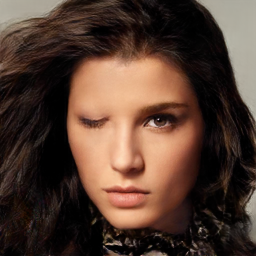}&
\includegraphics[width=0.35\linewidth]{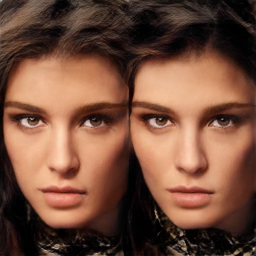}& \cna\\
\centertable{\rot{Ours}}&
\includegraphics[width=0.35\linewidth]{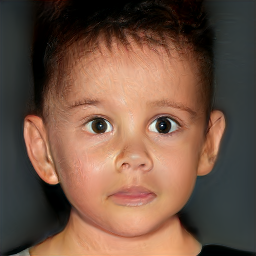}&
\includegraphics[width=0.35\linewidth]{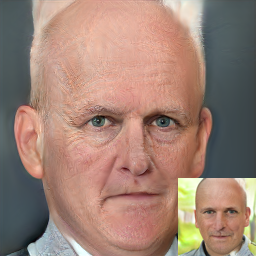}&
\includegraphics[width=0.35\linewidth]{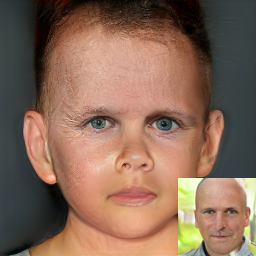}&
\includegraphics[width=0.35\linewidth]{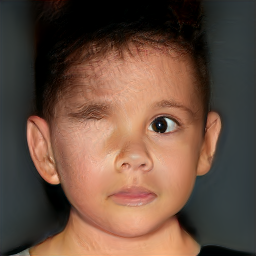}&
\includegraphics[width=0.35\linewidth]{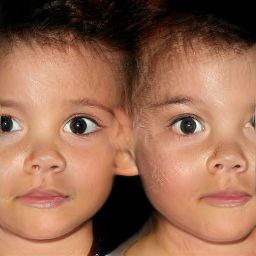}&
\includegraphics[width=0.35\linewidth]{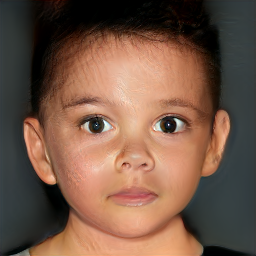}\\
%\rule{0pt}{1pt}\\ % small space
 & Original & Global & Local & Removing & Adding & Moving\\
\end{tabular}}
% \vspace{-10pt}
\caption{\textbf{Supported editing features.} While all of the tested methods enable global editing, only some offer local editing of part appearance, and non of the GAN-based methods (top three) demonstrated adding new parts, removing, nor animating parts via control handles, as we provide.
}
\label{fig:functionality}
\end{figure}

\subsection{Editing Quality}
\label{sec:image_quality}
Our goal is to minimize the inevitable drop in image quality when enabling different levels of editing. 
For \textbf{global editing}, our $\text{FID}_\text{lerp}$ is close to that of all but \cite{kim2021exploiting}, which however does not support positional editing.

\parag{User study.} Because there is no established metric for \textbf{local editing} and the autoencoder methods do not provide FID scores, we additionally conducted a comparative user study to  \cite{zhang2018unsupervised} (best autoencoder) on CelebA and \cite{zhu2020sean} (most disentangled prior work) on CelebA-HQ. Participants were asked to "Choose the image (A or B) with higher face quality" and "regardless of image quality" choose the pair that better preserves "facial features", "identity", and "outline of the face and its parts". The results are summerized in Table~\ref{tab:supp_survey}.
Compared to \cite{zhang2018unsupervised}, who supports the same keypoint editing as we do, our image quality is preferred drastically, by 92.17\%.
Also compared to, \cite{zhu2020sean} which reaches a better FID score before editing and better preserve the outlines with explicit segmentation masks (ours preferred 33.91\%, vs. 46.96\%), our face quality is preferred (by 94.78\%) after local editing mouth/eyes, and the identity preservation is nearly the same (49.57\% voted equal quality). We conclude that our edits are preferred because their masks and embeddings can become incompatible when exchanged while our keypoints are trained to be disentangled. 
The details are in the supplemental. \looseness=-1

\begin{table}
\begin{center}
\resizebox{0.98\linewidth}{!}{%
\begin{tabular}{|l|c|c|c|c|}
\hline
Aspect & Method to compare & In favour of ours & In favour of others & Equal quality\\  \hline
Editing image quality & Zhang et al. \cite{zhang2018unsupervised} & \textbf{92.17}\% & 0.87\% & 6.96\%\\ 
Editing image quality & SEAN \cite{zhu2020sean} & \textbf{94.78}\% & 2.61\% & 2.61\%\\ 
Part disentanglement & Zhang et al. \cite{zhang2018unsupervised} & \textbf{67.83}\% & 5.22\% & 26.95\%\\ 
Part disentanglement & SEAN \cite{zhu2020sean} & 28.69\% & 21.74\% & \textbf{49.57}\%\\ 
Identity preservation & Zhang et al. \cite{zhang2018unsupervised} & \textbf{55.65}\% & 14.78\% & 29.57\%\\ 
Shape preservation & SEAN \cite{zhu2020sean} & 33.91\% & \textbf{46.96}\% & 19.13\%\\ 
\hline
\end{tabular}}
\end{center}
\caption{\textbf{Survey results}.
% We our solution to \cite{zhang2018unsupervised} (autoencoder based) and \cite{zhu2020sean} (mask based) at hand of four different aspects: editing image quality, part disentanglement, shape preservation while changing expression, and shape preservation while changing appearance.
}
\label{tab:supp_survey}
\end{table}

For \textbf{image generation}, our results are in the same ballpark as the supervised image translation approaches (Table~\ref{tab:fid} FID, top half), yet above those unsupervised ones building upon StyleGAN (bottom half). This loosely quantifies the cost of imposing additional disentanglement constraints as similar training strategies and generator architectures are used.  \looseness=-1

\subsection{Disentangled Representations} \label{disentangled_representations}
The ability to move parts depends strongly on their disentanglement. 
We measure this as the overlap of the image regions controlled by pairs of parts using the CPD score explained in Figure~\ref{fig:CPD}. Table~\ref{tab:fid} compares CPD scores to all related methods that had trained models available. 
%\xz{not sure we still want to say overlap} it is still form of an overlap, just not a clean one
To compare to GANLocalEditing \cite{collins2020editing} with 8 parts, we grouped parts for SEAN and ours into semantically equivalent groups and, when comparing to GANLocalEditing, removed the background as not parametrized by them. %, including eyes, nose, mouth, and hair as individual parts. %
Our method scores the best across all unsupervised methods (0.63 CPD vs. 0.45, 0.39, 0.35) and matches that of the supervised SEAN on their labeled dataset (0.7 CPD vs. 0.7), despite them having the advantage of conditioning on sharp segmentation masks while we only condition on self-supervised keypoints. This demonstrates the improvement in part disentanglement brought about by our contributions.

\begin{table}
\begin{center}
\resizebox{\linewidth}{!}{%
\begin{tabular}{|l|l|c|c|c|c|}
\hline
\multicolumn{2}{|l|}{Method} & Conditioned on & FID $\downarrow$ & $\text{FID}_\text{lerp}$ $\downarrow$ & CPD $\uparrow$\\
\hline
\multicolumn{2}{|l|}{Pix2PixHD \cite{wang2018high}} & masks & 23.69$\star$ & - & -\\ 
\multicolumn{2}{|l|}{SPADE \cite{park2019semantic}} & masks & 22.43$\star$ & - & -\\ 
% SEAN \cite{zhu2020sean} & segmentation masks \& image & 17.66 & 30.29$\dagger$ (w/o bg 30.96) & 0.1312 (w/o bg 0.1640)\\
% Ours (with detection, like SEAN) & self-supervised keypoints, see Sec.~\ref{sec:unsupervised_keypoints_discovery} & 21.71 & 32.93 (w/o bg 28.85) & 0.1301 (w/o bg 0.1836)\\  \hline
\multicolumn{2}{|l|}{SEAN \cite{zhu2020sean}} & masks \& image & \textbf{17.66} & \textbf{30.29}$\dagger$ & \textbf{0.70}\\
\multicolumn{2}{|l|}{Ours (+self-sup. detector, see Sec.~\ref{sec:unsupervised_keypoints_discovery})} & image & 21.71 & 32.93 & \textbf{0.70}\\  \hline
\multicolumn{2}{|l|}{StyleGAN2 \cite{karras2020analyzing}} & unsupervised & 4.97 & 30.30$\dagger$ & -\\
\multicolumn{2}{|l|}{Alharbi et al. \cite{alharbi2020disentangled}} & unsupervised & - & 27.96$\dagger$ & -\\
\multicolumn{2}{|l|}{StyleMapGAN \cite{kim2021exploiting}} & unsupervised & \textbf{4.72} & \textbf{9.97} & -\\
GANLocalEditing & ($\epsilon=5$) & unsupervised & - & - & \textit{0.45}\\
Collins et al. \cite{collins2020editing} & ($\epsilon=50$) & unsupervised & - & - & \textit{0.39}\\
& ($\epsilon=500$) & unsupervised & - & - & \textit{0.35}\\
\multicolumn{2}{|l|}{Ours} & unsupervised & 16.26  & 31.80 & \textbf{\textit{0.63}}\\
\hline
% Ours (50k) &unsupervised, part-based & 10.66\\
% StyleGAN \cite{karras2019style} & part entangled & 5.06\\ \hline
% \cite{}
\end{tabular}}
\end{center}
\caption{\textbf{Portrait image generation and editing quality} 
% compared to supervised image translation on CelebA-HQ (top half) and to GANs on FFHQ (bottom). Our image quality scores before (FID column) and after editing  ($\text{FID}_\text{lerp}$) are close to the methods in each category, only outperformed by those providing fewer editing options or using supervision. This editing is enabled by the improved disentanglement (CPD column).
% $\star$ is trained by \cite{zhu2020sean} and $\dagger$ by \cite{kim2021exploiting} using official implementations.
}
\label{tab:fid}
\end{table}

\subsection{Generalization to Diverse Datasets}

\begin{figure*}[t]
\begin{center}
   \includegraphics[width=0.98\linewidth]{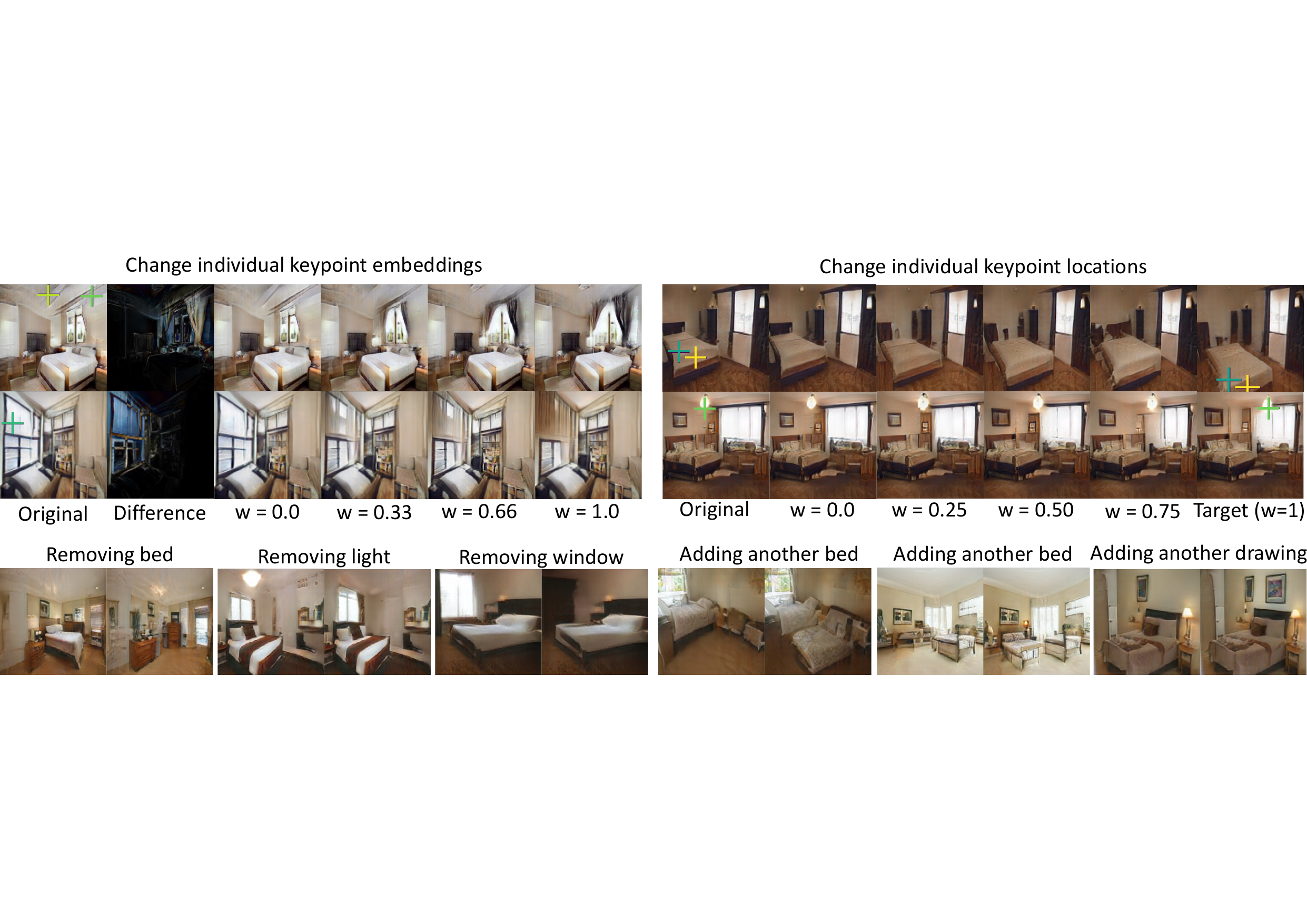}
%   \vspace{-10pt}
\end{center}
   \caption{\textbf{Editing on Bedroom} by (top-left) interpolating the keypoint embeddings of curtain and window and (top-right) moving bed and light; and (bottom) removing and adding objects}
\label{fig:bedroom}
\end{figure*}

\parag{LSUN Bedroom.} In Figure~\ref{fig:bedroom}, we explore the editing ability of entire scenes on the LSUN bedroom dataset. No previous unsupervised keypoint-based model has tried this difficult task before. We successfully interpolate the local appearance by changing the corresponding keypoint embeddings and translating the local key parts (window, bed) by moving the corresponding keypoints. Only \cite{collins2020editing} can change the appearance of parts, but they cannot move, remove or add individual parts; operations we support.

% \HR{This should go to limitations!}Since our learned representation is two-dimensional, it is not possible to rotate objects entirely as in recent 3D methods~\cite{3D NeRF GANs}.

\begin{figure*}[t]
\begin{center}
   \includegraphics[width=0.98\linewidth]{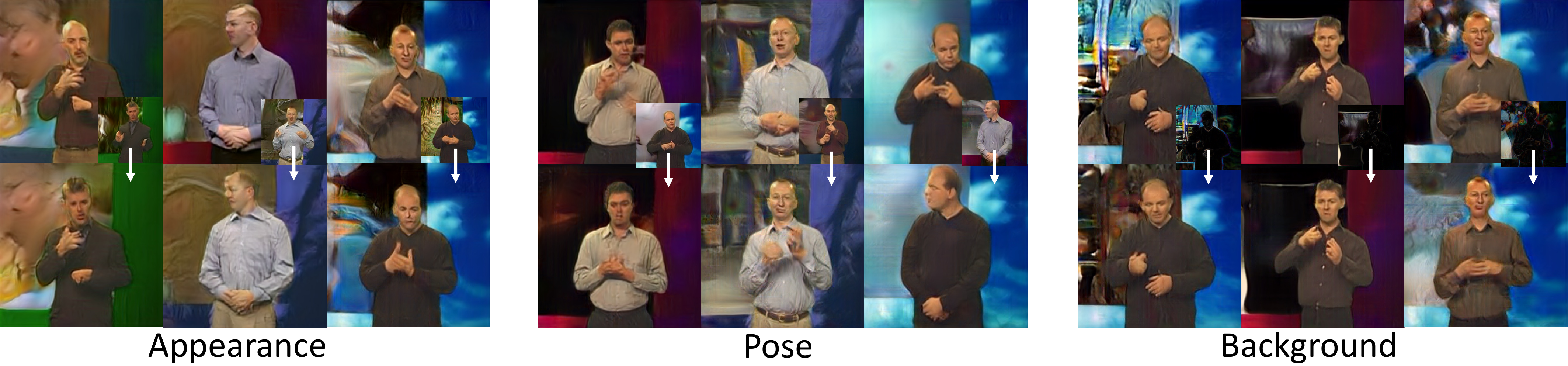}
%   \vspace{-10pt}
\end{center}
   \caption{\textbf{Editing on BBC Pose.} The first row shows the source image and the second row the editing results. 
   \textbf{Left:} the human appearance is swapped with the small target image. 
   \textbf{Center:} changing the position to the one in the overlay. 
   \textbf{Right:} changing the background (inset shows the difference). 
   }
\label{fig:bbcpose}
\end{figure*}

\textbf{BBC Pose.} Figure~\ref{fig:bbcpose} explores the editing of persons. Although artifacts remain due to the detailed background and motion blur in the datasets, pose and appearance can still be exchanged.   \looseness=-1

\subsection{Ablation Tests}

We demonstrate the importance of our contributions by (1) removing the background; (2) removing the global style vector; (3) using additive global style vector instead of multiplicative ones; (4) using contrastive learned keypoint embeddings instead of multiplicative ones; (5) removing the keypoint embedding; (6) removing keypoints.  Table \ref{tab:ablation_archi} shows that all additions increase one or more of the keypoint localization ($L_2$  error normalized by interocular distance), FID, or CPD scores. 
To speedup experimentation, the FID is calculated on FFHQ at resolution $256\times256$,
%between 50k generated images and the original dataset, 
and CPD is calculated on all 10 keypoints with the background included. 

\begin{figure}[t!]
\begin{center}
\includegraphics[width=0.99\linewidth]{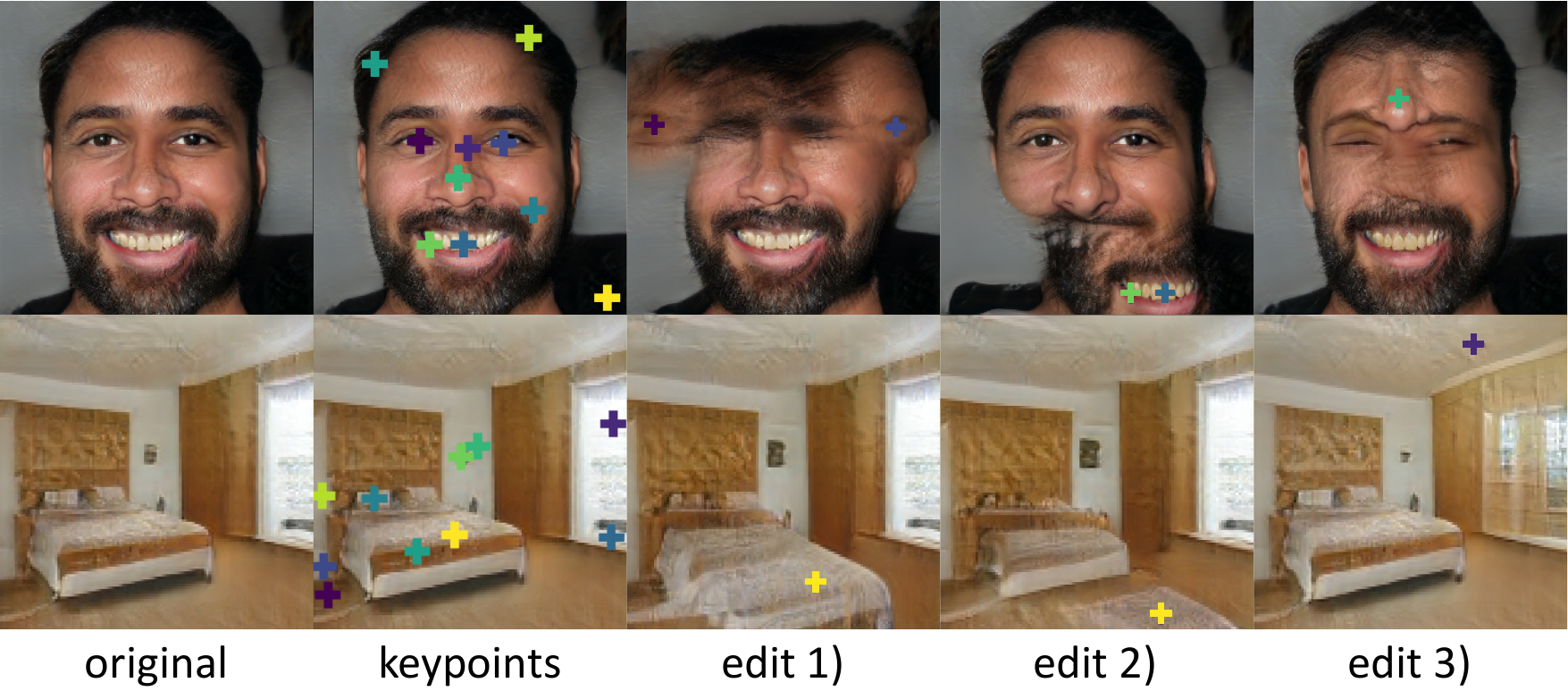}
\end{center}
  \caption{\textbf{Out-of-distribution positions}. In the first row, 1) the eyes disappear when the keypoints are moved outside the facial region; 2) the mouth is maintained even if the keypoints are far from the original; 3) notably, the nose retains the shape when moved to the forehead while not influencing the neighboring eyebrows. In the second row, 1,2): moving one of the bed keypoints first enlarges the bed and then breaks it into two pieces; 3) the window disappears when the keypoint moves to the ceiling.}
\label{fig:extreme_pos}
\end{figure}

We also include examples in Figure~\ref{fig:extreme_pos}, where keypoints are moved to out-of-distribution positions. Some of the parts become invisible or distorted in unrealistic positions but the remaining parts are intact and interact in a plausible way with nearby parts. For instance, the nose can be moved up as one part and connects to the eyebrows without distorting them. This highlights the strong disentanglement of parts and supports the numerical analysis in Table~\ref{tab:fid}. Notably only the relative positioning of parts influences the image, without having any bias to the absolute image center position, as shown by moving all keypoints in the teaser Figure~\ref{fig:teaser}.

\begin{table}
 \vspace{-0pt}
 \begin{center}
\resizebox{0.98\linewidth}{!}{%
\begin{tabular}{|l|c|c|c|}
\hline
Method & relative $L_2$ error \% $\downarrow$ & FID $\downarrow$ & CPD $\uparrow$ \\ 
\hline
adding global style vector & 5.29\% & 42.02 & 0.72\\ 
w/o keypoint & - & 34.69 & -\\ 
w/o keypoint embedding & 22.81\% & 32.41 & -\\ 
w/o global style vector & 6.76\% & 28.75 & -\\ 
contrastive keypoint embedding & 7.53\% & 28.47 & 0.76\\
w/o background & 6.43\% & 25.67 & 0.71\\ 
full model & 5.85\% & 23.50 & 0.67\\ 
\hline
\end{tabular}}
\end{center}
% \vspace{-5pt}
\caption{\textbf{Quantitative ablation on keypoint localization} (
% $L_2$  error normalized by interocular distance), image quality (FID), and Correlation-based Part Disentanglement (CPD), with all contributions improving either one or more of the metrics.
}
\label{tab:ablation_archi}%
\end{table}%

\subsection{Keypoint Consistency and Interpretability} \label{sec:unsupervised_keypoints_discovery}
To be interpretable, the learned keypoints must correspond to the same semantic image region across different images. We test this with the detector trained on our generated examples (see variants paragraph), which can only succeed when parts are placed consistently.
For evaluation, we follow \cite{thewlis2017unsupervised} and subsequent methods: As the order and semantics of unsupervised keypoints are undefined, a linear regressor from the predicted keypoints to the 5 ground truth keypoints is learned on the MAFL training set~\cite{zhang2014facial}.
%, zhang2018unsupervised, jakab2018unsupervised, lorenz2019unsupervised
% All images are resized to $128\times 128$ and the training and test set of the MAFL subset from CelebA are excluded when training the LatentKeypointGAN. 
The test error is the $L_2$ error normalized by the inter-ocular distance.
High accuracy can therefore only be obtained when the learned disentangled keypoints move consistently with the human-annotated ones. 
% architecture without adapting to this new task

LatentKeypointGAN strikes a low error of 5.9\%, which lies between 8.0\% by \cite{thewlis2017unsupervised} and 3.2\% by \cite{jakab2018unsupervised}, validating the interpretability and consistently of the learned keypoints. Furthermore, LatentKeypointGAN-tuned variant reaches 3.3\%, thereby contesting existing unsupervised keypoint detection approaches with an alternative methodology using GANs.
\section{Limitations and Future Work}
\label{sec:limitations}
Our model brings a new trade-off between the disentanglement required for editing, generality to diverse domains, and the image quality. 
While successful in disentangling facial details and objects in the bedroom dataset, the hair in portraits can mix with the background and, on the other hand, locally encoded features can lead to asymmetric faces, such as a pair of glasses with differently styled sides. 
% If a different level of entanglement is desired, one could attempt to tune $\tau$ for each application domain or even each keypoint.
For BBC Pose, the keypoints are not consistent, which could be overcome by linking keypoints with a skeleton.
While the face orientation in portrait images can be controlled by moving keypoints, we found that orientation changes on the bedroom images are not reliable. We believe that it will be necessary to learn a 3D representation and datasets with less-biased viewpoints. 

\section{Conclusion}

We present a GAN-based framework that is internally conditioned on keypoints and their appearance encoding. By learning disentangled representations from scratch instead of starting from a pre-trained StyleGAN, we provide an interpretable hidden space that enables intuitive spatial editing via control handles. 
%We provide the idea and implementation to avoid issues with TPS and related deformation models that are common in existing autoencoders. 
This LatentKeypointGAN also facilitates the generation of image-keypoint pairs, thereby providing a new methodology for unsupervised keypoint detection that is typically addressed with autoencoders. 
% local editing without using any comparison of pair of images
%We argue that our model has comparative or even larger potential than paired models.  We expect that researchers will push GAN and spatially adaptive image-to-image translation even further. Then we can further improve image generation and keypoint localization of LatentKeypointGAN.

\bibliographystyle{IEEEtran}
\bibliography{sn-bibliography}

\appendix
\section*{}
In this supplemental document, we present additional details, such as on the new CPD metric, neural network architectures, the progressive training, and hyperparameters. Furthermore, we added hundreds of additional qualitative results in the additional\_results.PDF file and animation examples in the supplemental videos. 
\textbf{\href{https://xingzhehe.github.io/LatentKeypointGAN/}{project website}}.

\section{More Related Work} \label{sec:related_work}
\textbf{Editing entire images with GANs.}
More recently, efforts have been made on exploring the latent space of a pre-trained StyleGAN for image editing~\cite{shen2020interpreting, Jahanian*2020On}. To allow editing real-world images, various encoders~\cite{zhu2020domain, abdal2019image2stylegan, abdal2020image2stylegan++, guan2020collaborative, wulff2020improving, richardson2020encoding} have been trained to project images into the latent space of StyleGANs. These methods provide control over the image synthesis process, such as for changing age, pose, gender, illumination and enable rig-like controls over semantic face parameters~\cite{tewari2020stylerig, tewari2020pie, ghosh2020gif, deng2020disentangled} by conditioning on parametric face models~\cite{blanz1999morphable, FLAME:SiggraphAsia2017}. 
Compared with these methods, our model focuses on detailed and local semantic controls. Instead of changing the face as a whole, our method is able to change a local patch without an unwanted impact on other regions. Furthermore, our keypoints provide control handles for animation without manual rigging, making it easily applicable to different objects and image domains. 

\textbf{Conditioned GAN} usually synthesize images that resemble a given reference input, including segmentation masks~\cite{sola2017image,zhu2017unpaired,park2019semantic}, scene graphs~\cite{johnson2018image, li2021controllable}, surface normals~\cite{wang2016generative}, part labels~\cite{cheng2020segvae}, and human pose~\cite{men2020controllable}. They achieve local editing by modifying the explicit part labels~\cite{johnson2018image, li2021controllable, cheng2020segvae} or the latent codes for a mask segment~\cite{park2019semantic, men2020controllable}. The closest to ours are the mask-conditioned image synthesis methods.
\cite{park2019semantic} pioneered using spatially-adaptive denormalization (SPADE) to transfer segmentation masks to images, which we borrow and adapt to be conditioned on landmark position. To control individual aspects of faces, such as changing eye or nose shape, recent works~\cite{zhu2020sean, tan2021efficient, tan2021diverse}
% tan2020michigan
further modify SPADE to allow local editing on images. 
The recently proposed DatasetGAN \cite{zhang2021datasetgan} and its application EditGAN \cite{ling2021editgan} significantly reduce the required number of annotated examples to less than a hundred.
Compared with these methods, our model does not take any kind of supervision or other conditions at training time. It is trained in a totally unsupervised manner and therefore eases the application to new domains for which no labels exist. Still, our method allows the landmarks to learn a meaningful location and semantic embedding that can be controlled at test time. 

\textbf{Unsupervised landmark discovery} methods aim to detect the landmarks from images without manual labels. Most works train two-branch autoencoders, where shape and appearance are disentangled by training on pairs of images where one of the two is matching while the other factor varies. The essence of these methods is to compare the pair of images to discover disentangled landmarks. These pairs can stem from different views~\cite{suwajanakorn2018discovery, Rhodin_2019_CVPR} and frames of the same video~\cite{siarohin2019animating,siarohin2019first,siarohin2021motion,wang2021one} in almost static backgrounds. However, this additional motion information is not always available or is difficult to capture. Existing unsupervised methods trained on single images create pairs from spatial deformation and color shifts of a source image~\cite{xing2018deformable,shu2018deforming,thewlis2019unsupervised,li2020unsupervised,cheng2020unsupervised,dundar2020unsupervised}. Yet, the parameters of augmentation strategies, such as the commonly used thin-plate spline deformation model~\cite{thewlis2017unsupervised, zhang2018unsupervised,jakab2018unsupervised,lorenz2019unsupervised}, are difficult to calibrate, and image quality is lower than for GANs.
Moreover, the cycle consistency-based methods~\cite{wu2019transgaga, xu2020unsupervised} cannot provide local editing ability. 
By contrast, we show that our generator can all-together disentangle the appearance and keypoint location, control the images locally, and provides an improved image quality. 
Part segmentation methods can also provide keypoints, defined at the center of each mask, but they usually require weak supervision on saliency maps \cite{hung2019scops, NEURIPS2021_ec8ce6ab}. 
Liu et al.~\cite{liu2021unsupervised} remove the requirement of saliency maps, but assume the object to be centered in the image, which is constraining.
Efforts on unsupervised 3D keypoint detection~\cite{fernandez2020unsupervised, jakab2021keypointdeformer} have been made on 3D shapes by exploring the symmetry~\cite{fernandez2020unsupervised} and by interpolated deformation from one shape to another~\cite{jakab2021keypointdeformer}. 
Unlike them, we focus on 2D keypoints in images.
% \HR{We may want to remove or shorten this paragraph significantly} \xz{How about removing it? I think SIGGRAPH does not care too much at this part?}\HR{If so, we should make sure that we cite the most important works in the intro (where we discuss the autoencoder work} \xz{I have cited the most important works in the intro. Just wondering if we could move this part to appendix?}

\textbf{Multi-stage trained GANs.} Researchers sometimes divide the training of GANs into different stages, where different intermediate representations are usually learned for better performance or better controllability. The representations include surface normals~\cite{wang2016generative}, scene layouts~\cite{johnson2018image, li2021controllable}, or segmentation masks~\cite{men2020controllable}. While they require explicit supervision of the ground truth annotation, Karras et al.~\cite{karras2018progressive, karras2019style} proposed using smaller images as the intermediate representations. They progressively increase the image resolution in each stage to generate high-resolution high-quality images, which is also adopted in our network training.

\textbf{Diffusion models} \cite{ho2020denoising, dhariwal2021diffusion} nowadays can generate high quality images by denoising a Gaussian noise map. Benefited from the large-scale text-image models \cite{radford2021learning}, large diffusion models can generate various images by using text \cite{ramesh2021zero, saharia2022photorealistic, rombach2022high}. Zhang and Agrawala \cite{zhang2023adding} proposed to ControlNet, injecting spatial conditions to the feature maps of the pre-trained diffusion models to control the spatial information of the generated images. We share the similar purposes but our condition, keypoint, itself is learned along with the generator. Furthermore, our learned keypoints can be further fed into ControlNet to control the diffusion models.
\section{Correlation-based Part Disentanglement} \label{supp:cpd}
Disentanglement is measured as the linear correlation betwenn editing operations on the image. Let $\mI_\text{original},\allowbreak \mI_\text{target} \in \myR^{H\times W\times3}$ be the original and target image, respectively. Denote $\mI_k \in \myR^{H\times W\times3}$ be the edited image with part embedding $\vw_k$ from $\mI_\text{target}$ and other part embeddings from $\mI_\text{original}$. We take $L_2$-norm for each pixel of the difference map of $\mI_k$ and $\mI_\text{original}$, resulting a difference heatmap
\begin{equation}
    \mD_k(\vp) = \|\mI_k(\vp)-\mI_\text{original}(\vp)\|_2     \in \myR^{H \times W}
\label{cpe:diff_map}
\end{equation}
where $\vp$ denote an arbitrary pixel. We now flatten $\mD_k$ into a vector $\vd_k \in \myR^{H W}$ to calculate the correlation matrix $\rho\in\myR^{K\times K}$ for all pairs $i,j$ of the $K$ parts,
\begin{equation}
    \rho(i, j) = \frac{\vd_i \cdot \vd_j}{\|\vd_i\|_2\|\vd_j\|_2}.
\end{equation}
We calculate the correlation matrices for $N=2000$ images, and then use the average of the off-diagonal elements of the averaged correlation matrices as the part disentanglement score,
\begin{equation}
    \text{CPD} = 1- \sum_{i\neq j}\frac{1}{N K(K-1)}\sum_{n=1}^N\rho_n(i,j),
\end{equation}
where the $1-\sum$ ensured that this score goes from 0 to 1, with 1 representing perfect disentanglement.

 \begin{figure*}[t]
   \resizebox{0.98\linewidth}{!}{%
 \begin{tabular}{cccc}
 \includegraphics[width=0.5\linewidth]{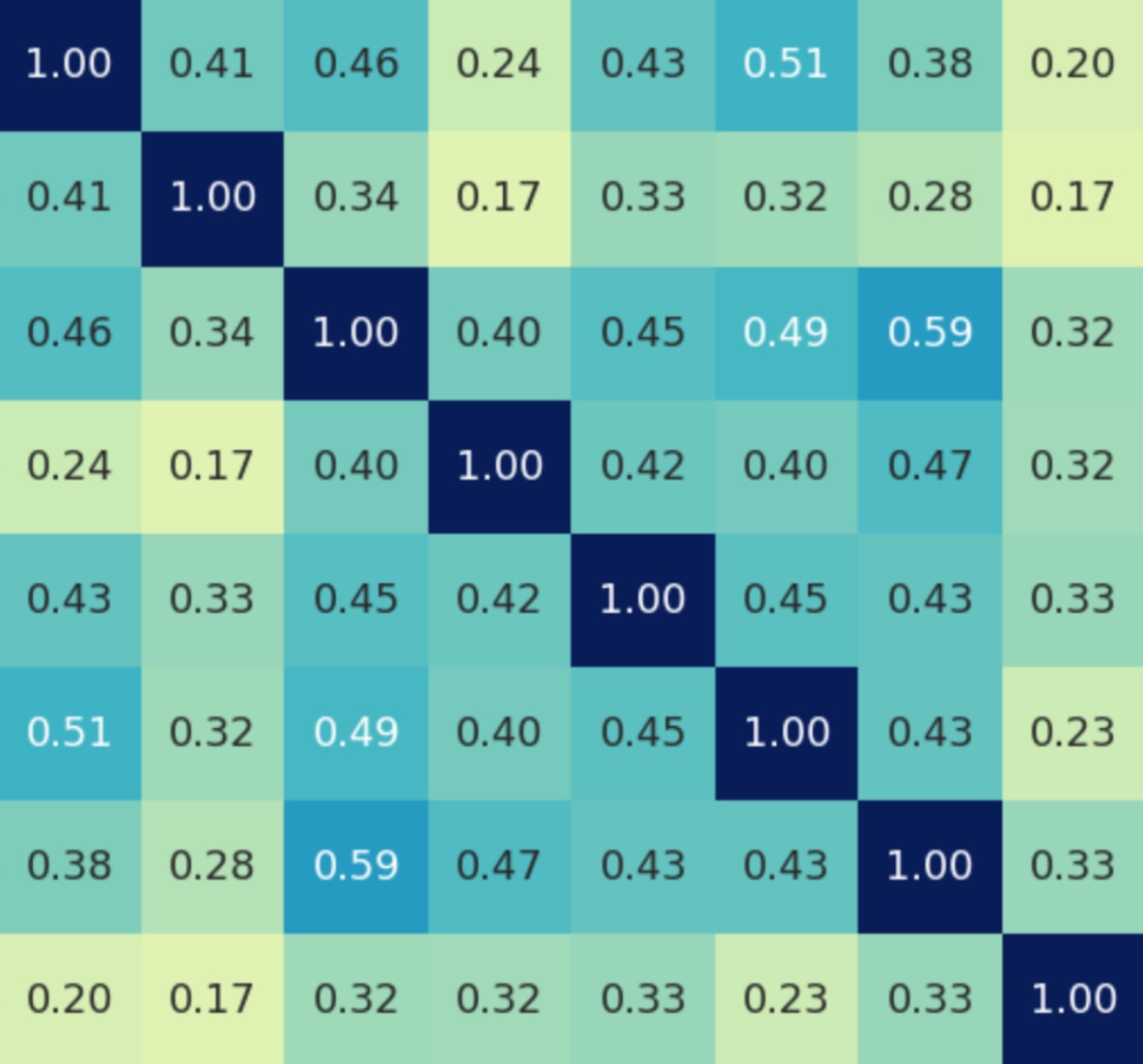}&
 \includegraphics[width=0.5\linewidth]{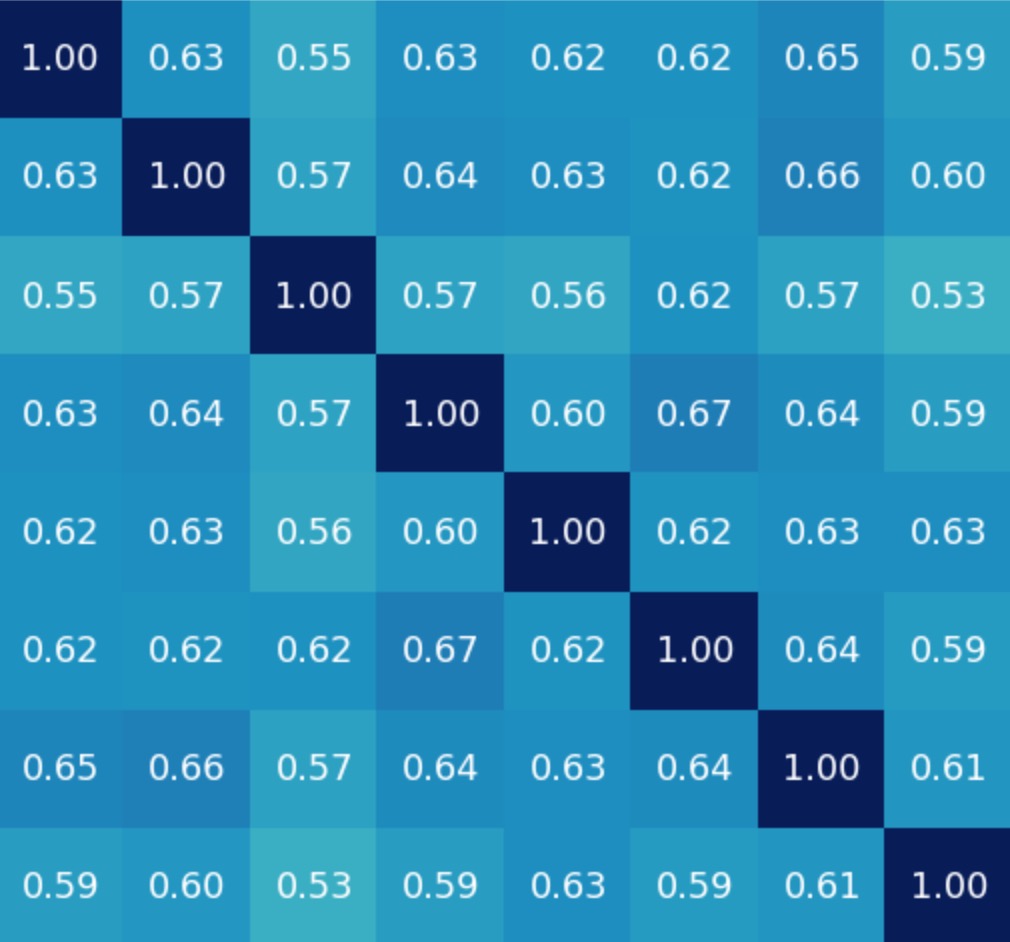}&
 \includegraphics[width=0.5\linewidth]{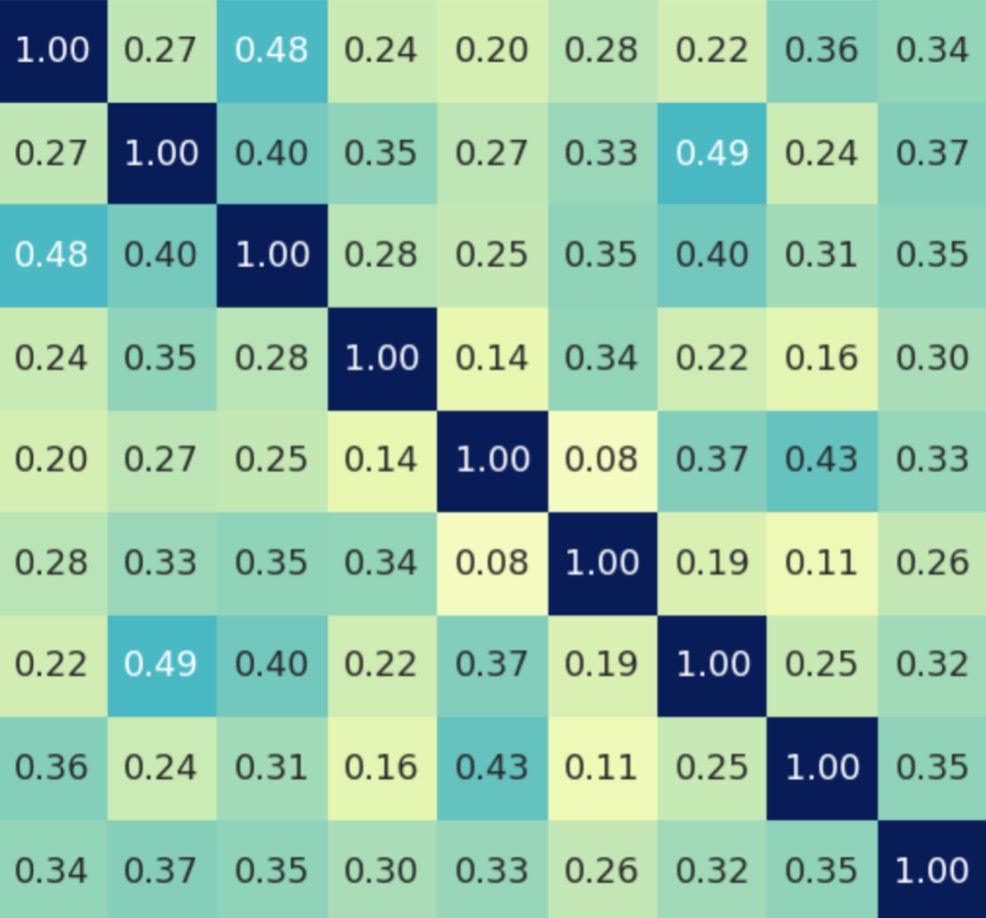}&
 \includegraphics[width=0.5\linewidth]{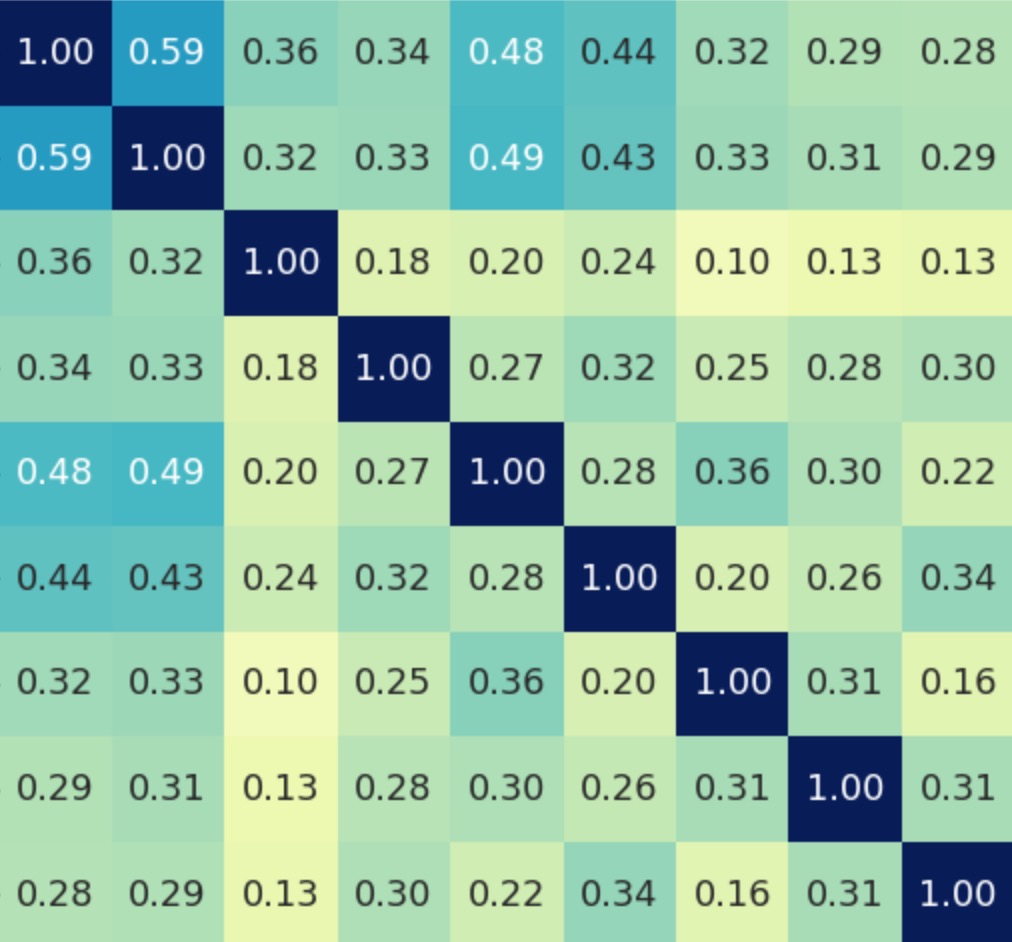}\\
 \rule{0pt}{1pt}\\ % small space
 Ours (FFHQ)  & \cite{collins2020editing} (FFHQ) & Ours (CelebA-HQ) & \cite{zhu2020sean} (CelebA-HQ)\\
 \end{tabular}}
 \caption{\textbf{Disentanglement comparison using CPD.} We plot the correlation matrix to visualize the part distanglement for SEAN \cite{zhu2020sean}, GANLocalEditing \cite{collins2020editing} and ours. Note that on FFHQ we do not include background to make a fair comparison with GANLocalEditing.
 }
 \label{fig:correlation}
 \end{figure*}

Figure~\ref{fig:correlation} visualize the resulting part correlation matrices used for CPD, our approach shows less correlation between parts (brighter off-diagonal elements). We largely outperfrom the GAN-based approach from \cite{collins2020editing}, showing significantly less correlation while matching to that of \cite{zhu2020sean}, who however use segmentation masks as additional cues.

% \parag{Correlation-based Part Expressivity (CPE).} We first use Equation~\ref{cpe:diff_map} to obtain the difference map and reshape it to a vector $\vd_k$. Then we calculate the expressivity $r$ for all parts by
% \begin{equation}
%     r = \frac{\sum_{k=1}^K\|\vd_k\|_2}{\|\mI_\text{original} - \mI_\text{target}\|_2},
% \end{equation}
% where $K$ is the number of parts. The CPE is obtained by averaging on $N$ different samples,
% \begin{equation}
%     \text{CPE} = \frac{1}{N}\sum_{n=1}^N r_n.
% \end{equation}
\section{Keypoint Consistency} 
\begin{figure*}[ht]
\begin{center}
   \includegraphics[width=0.98\linewidth]{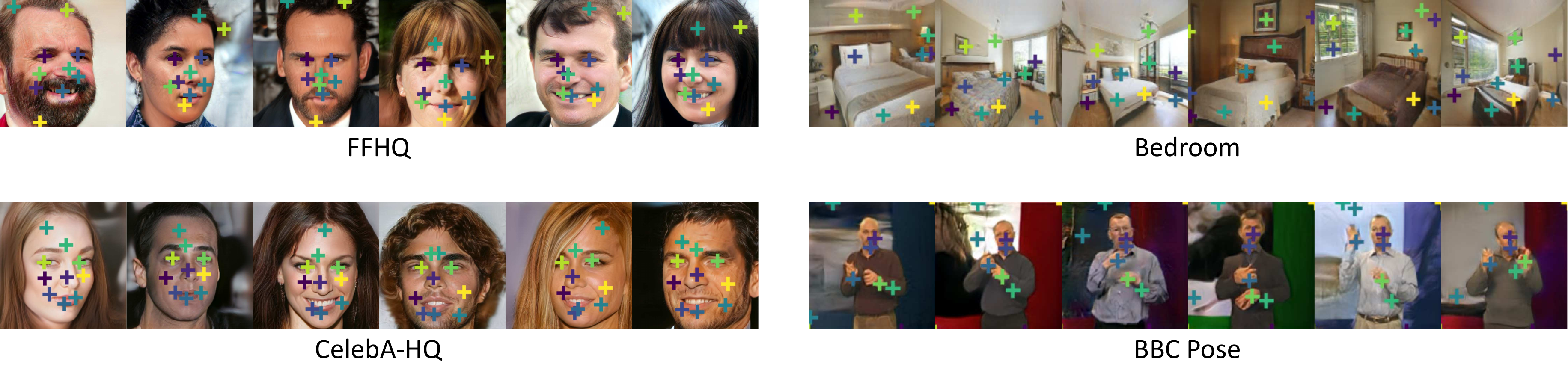}
\end{center}
   \caption{\textbf{Keypoints.} We show the keypoints on each dataset.}
\label{fig:keypoints}
\end{figure*}

Figure~\ref{fig:keypoints} shows the keypoints generated on all datasets. The keypoints are semantically meaningful and very consistent across different instances.

\section{Disentangled Background}
\begin{figure}
 \vspace{1pt}
  \includegraphics[width=1\linewidth]{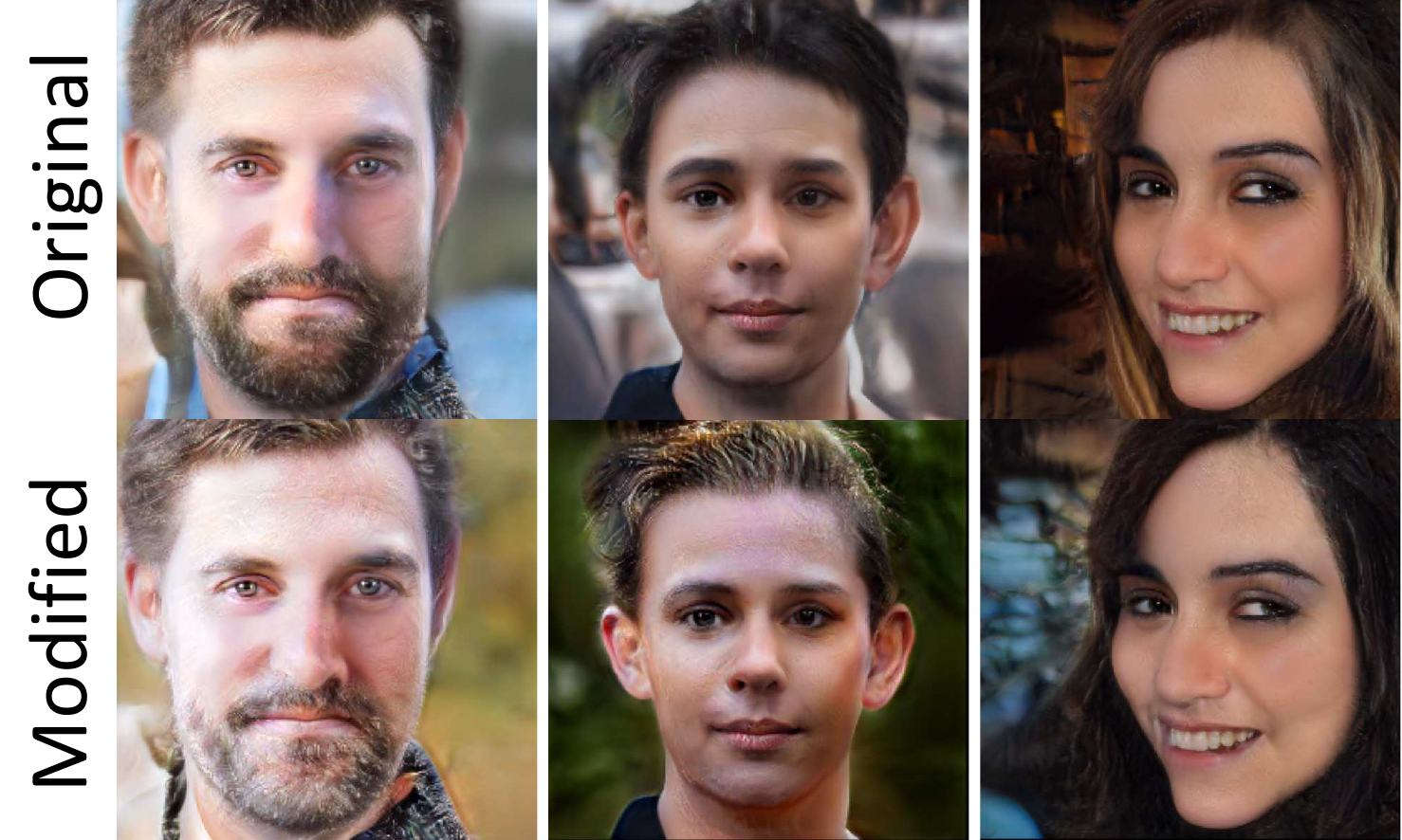}
  \caption{\textbf{Disentangled Background.} The background is changed while the faces are fixed.}
\label{fig:bg_dis}
\end{figure}

Figure \ref{fig:bg_dis} shows a faithful transition between backgrounds while keeping the face fixed. To this end, we fix the keypoint noise $\vz_\text{kp\_pose},\allowbreak \vz_\text{kp\_app}$, and change only the background noise input, $\vz_\text{bg\_emb}$. The local change in the three diverse examples shows that the background and keypoint encodings are disentangled well. Note however that the illumination and hair color is learned to be part of the background, which makes some sense as these global feature cannot be attributed to individual keypoints.

\section{Tuning LatentKeypointGAN for Keypoints Detection}  \label{sec:supp_detection}

We desire an architecture that encodes the location of parts solely in the keypoint locations to improve keypoint localization and the subsequent learning of a detector. Even though the convolutional generator is translation invariant, additional care is necessary to prevent leakage of global position at the image boundary and from the starting tensor. All these tuning steps are explained below.

\paragraph{Padding Margins.} As pointed out by \cite{islam2020much, alsallakh2021mind, xu2020positional, kayhan2020translation}, convolutions with zero padding are very good at implicitly encoding absolute grid position at deeper layers. To prevent this, we follow \cite{karras2021alias}. By maintaining a fixed margin around the feature map and cropping after each upsampling, we effectively prevent the leaking of absolute position and the bias to the center because none of the generated pixels ever reaches a boundary condition. We use a 10-pixel margin. Note that such measures do not directly apply to autoencoders who are bound to the fixed resolution of the input.

\paragraph{Positional Encoded Starting Tensor.} We remove the $4\times4\times512$ starting tensor because it can encode absolute grid position. We replace it with the positional encoding $\mM$ of difference between keypoints $\vk_1,...,\vk_K$ and the grid positions $\vp$, 
\begin{equation}
    \begin{aligned}
        \mM(\vp) = [&\sin(\pi*\text{Linear}([\vp-\vk_1, ..., \vp-\vk_K])), \\
        &\cos(\pi*\text{Linear}([\vp-\vk_1, ..., \vp-\vk_K]))].
    \end{aligned}
\end{equation}
The underlying idea is to only encode relative distances to keypoints but not to the image boundary.

\paragraph{Larger Starting Tensor.} We found that starting the feature map from $32\times32$ instead of $4\times4$ improves keypoint localization accuracy. 
%Note that the starting feature map is not the the intermediate first feature map that is supervised to generate RGB images, whose resolution is still $64\times64$.

\paragraph{Background Handling.} The background should be consistent across the whole image, but complete occlusion is inaccurate to model with the sparse keypoints and their limited support in the intermediate feature maps. Hence, we introduce the explicit notion of a foreground mask that blends in the background. The mask is generated as one additional layer in the image generator. To generate the background, we use a separate network that is of equivalent architecture to LatentKeypointGAN-tuned. Note that in background generation we use AdaIN \cite{huang2017arbitrary} instead of SPADE because there is no spatial sensitive representation, such as keypoints. Foreground and background are then blended linearly based on the foreground mask. We use an $1\times1$ convolution layer to generate the final RGB images.

\paragraph{Simplification} In addition, we remove the progressive training \cite{karras2018progressive} and use the ResBlocks as defined by \cite{park2019semantic}. This step is for simplification (quicker training results on medium-sized images) and does not provide a measurable change in keypoint accuracy.

As shown in Table~\ref{tab:supp_keypoint_detection}, each step of improvement contributes significantly to the keypoint accuracy and consistent improvement on prior works on the two in-the-wild settings. Please see the main paper for the complete comparison. The FID of LatentKeypointGAN-tuned on CelebA of resolution $128\times 128$ is 18.34. Because image quality is not significantly improved, we keep the simpler LatentKeypointGAN for editing and the tuned version only for keypoint detection.

\begin{table*}[t]
\centering
% \resizebox{0.98\linewidth}{!}{%
\begin{tabular}{|l|c|c|c|}
\hline
Method & Aligned (K=10) & Wild (K=4) & Wild (K=8)  \\ \hline
\cite{thewlis2017unsupervised} & 7.95\% & - & 31.30\% \\ 
\cite{zhang2018unsupervised} & 3.46\% & - & 40.82\% \\ 
\cite{lorenz2019unsupervised} & 3.24\% & 15.49\% & 11.41\% \\
IMM \cite{jakab2018unsupervised} &  3.19\% & 19.42\% & 8.74\% \\
\cite{dundar2020unsupervised} &  \textbf{2.76}\% & - & - \\ \hline
LatentKeypointGan-tuned & 3.31\% & \textbf{12.1}\% & \textbf{5.63}\%\\ 
- larger starting tensor & 3.50\% & 14.22\% & 8.51\% \\
- separated background generation & 4.24\% & 19.29\% & 14.01\% \\
- positional encoded starting tensor & 5.26\% & 24.12\% & 23.86\% \\ 
- margin & 5.85\% & 25.81\% & 21.90\%
\\ \hline
\end{tabular}
\caption{\textbf{Landmark detection on CelebA (lower is better)}. The metric is the landmark regression (without bias) error in terms of mean $L_2$ distance normalized by inter-ocular distance. The bottom four rows shows our improvement step by step. We use the same number of keypoints as previous methods.} 
\label{tab:supp_keypoint_detection}
\end{table*}
\section{Image Editing Quality Comparison} \label{sec:supp_quality}
\begin{figure*}[h]
  \centering
%   \setlength{\tabcolsep}{0.5pt}
%   \renewcommand{\arraystretch}{0.1}
%   \small
  \resizebox{0.98\linewidth}{!}{%
\begin{tabular}{cccccccccccc}
\multicolumn{4}{c}{Comparison with mask-conditioned methods (CelebA-HQ)} & \multicolumn{1}{c}{}&  \multicolumn{3}{c}{Unsupervised keypoint-based methods (CelebA $128\times128$)} & \multicolumn{1}{c}{}& \multicolumn{2}{c}{Unsupervised GAN (FFHQ)}\\
\\
\includegraphics[width=0.25\linewidth]{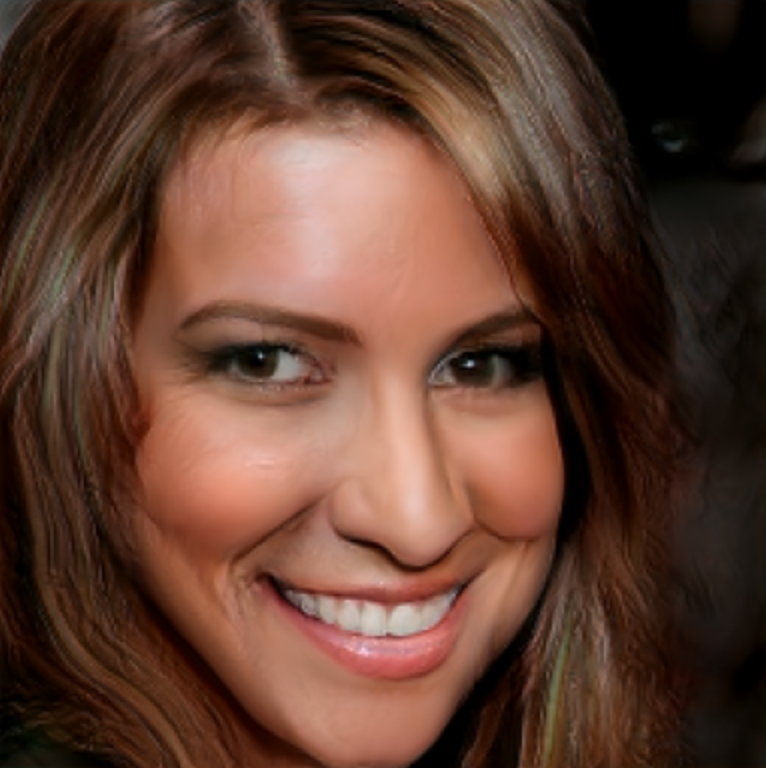}&
\includegraphics[width=0.25\linewidth]{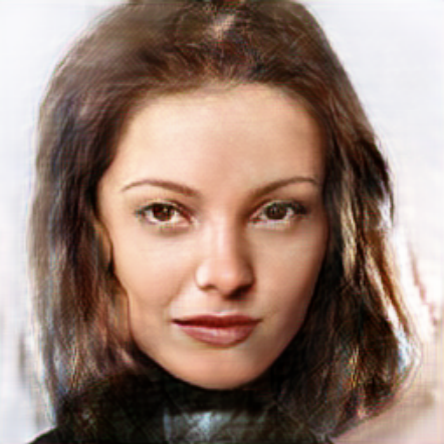}&
\includegraphics[width=0.25\linewidth]{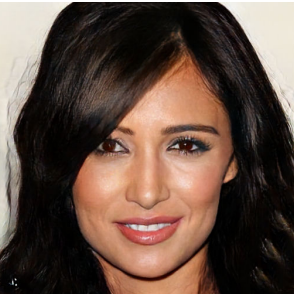}&
\includegraphics[width=0.25\linewidth]{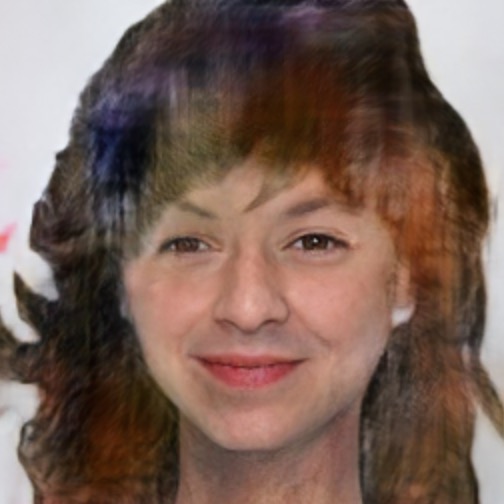}& 
\multicolumn{1}{c}{}&
\includegraphics[width=0.25\linewidth]{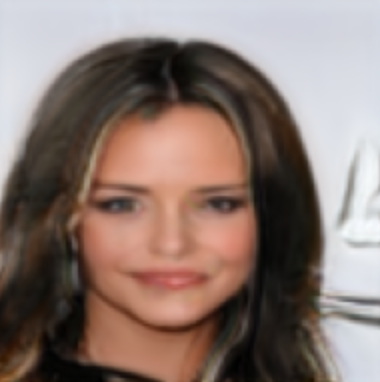}&
\includegraphics[width=0.25\linewidth]{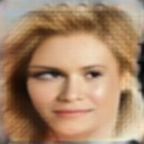}&
\includegraphics[width=0.25\linewidth]{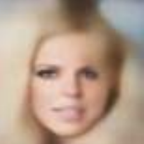}&
\multicolumn{1}{c}{}&
\includegraphics[width=0.25\linewidth]{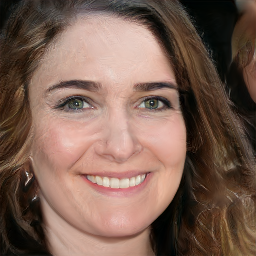}&
\includegraphics[width=0.25\linewidth]{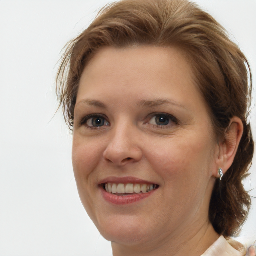}&\\

\includegraphics[width=0.25\linewidth]{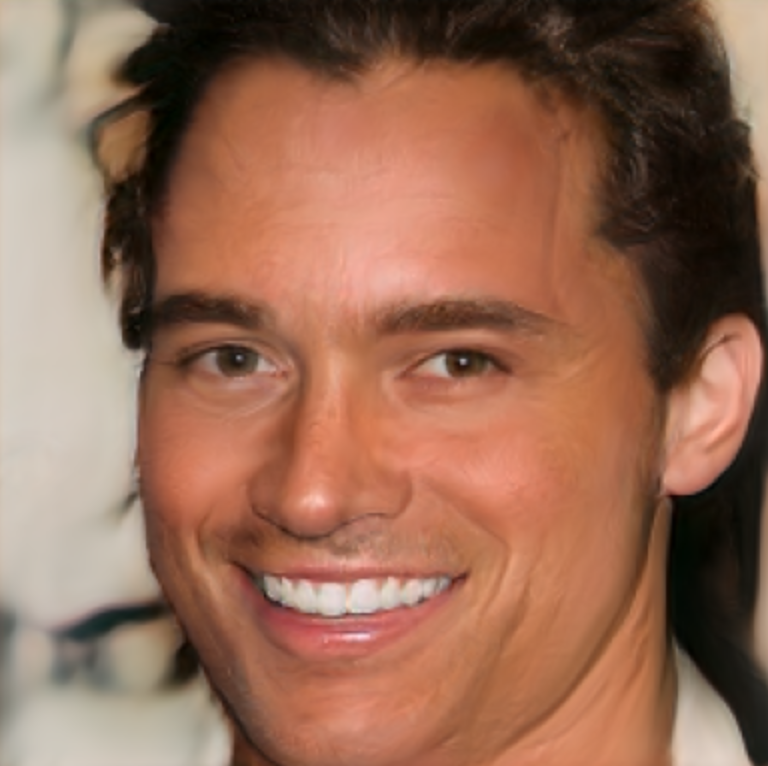}&
\includegraphics[width=0.25\linewidth]{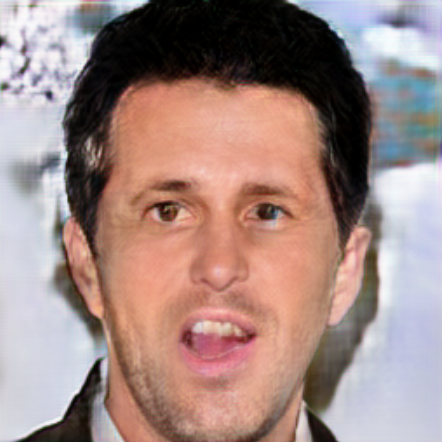}&
\includegraphics[width=0.25\linewidth]{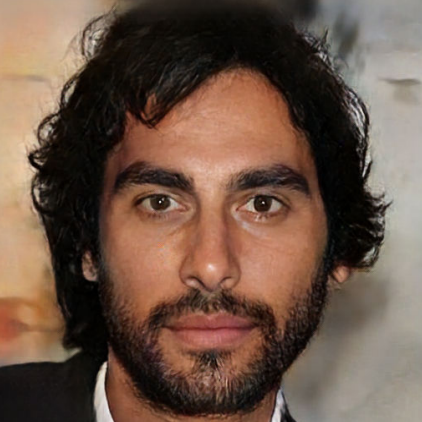}&
\includegraphics[width=0.25\linewidth]{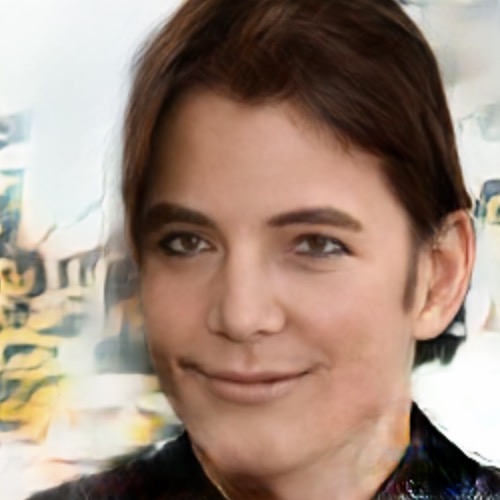}&
\multicolumn{1}{c}{}&
\includegraphics[width=0.25\linewidth]{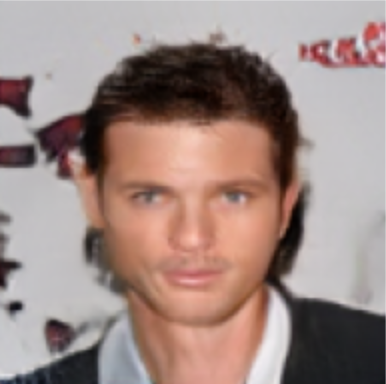}&
\includegraphics[width=0.25\linewidth]{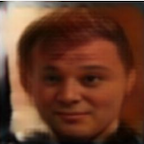}&
\includegraphics[width=0.25\linewidth]{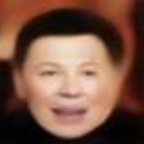}&
\multicolumn{1}{c}{}&
\includegraphics[width=0.25\linewidth]{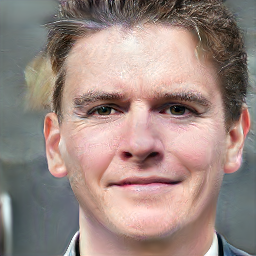}&
\includegraphics[width=0.25\linewidth]{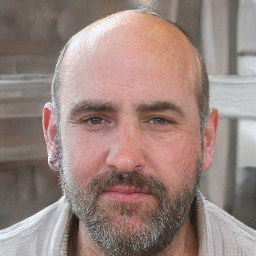}&\\
\rule{0pt}{1pt}\\ % small space
Ours & SPADE \cite{park2019semantic} & MaskGAN \cite{lee2020maskgan} & SEAN \cite{zhu2020sean} &\multicolumn{1}{c}{} & Ours & \cite{xu2020unsupervised} & \cite{zhang2018unsupervised} &\multicolumn{1}{c}{} & Ours & \cite{collins2020editing} &\\
\end{tabular}}

\caption{\textbf{Image editing quality comparison.} We compare the image editing quality with both, supervised (left) and unsupervised (middle, right). LatentKeypointGAN improves on the methods in both classes.}
\label{fig:comparison}
\end{figure*}

Figure~\ref{fig:comparison} validates the improved editing quality, showing comparative quality to conditional GAN (supervised by paired masks), and superior quality to unsupervised methods (greater detail in the hair and facial features). Note that for SEAN \cite{zhu2020sean} we use the edited images (combine mask with different face appearance) instead of reconstructed images (combine mask with the corresponding face appearance) for fair comparison with other GAN methods. The qualitative improvement is further estimated in the subsequent user study.

The main paper reports FID scores for all those datasets where prior work does. To aid future comparisons we also report the remaining FID scores: 17.88 on FFHQ, 18.25 on CelebA, 30.53 on BBC Pose, and 18.89 on LSUN Bedroom. The FID is calculated by the generated 50k images and the resized original dataset.

\begin{figure*}[t]
\begin{center}
   \includegraphics[width=0.98\linewidth]{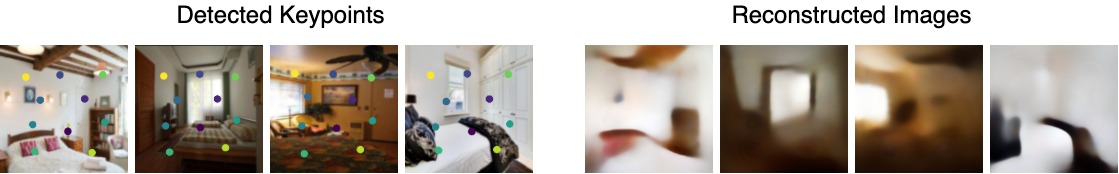}
\end{center}
   \caption{\textbf{\cite{lorenz2019unsupervised} on LSUN Bedroom.} (Left) Detected keypoints. The keypoints are static and do not have semantic meaning. (Right) Reconstructed images. The reconstruction completely fails.}
\label{fig:lorenz_bedroom}
\end{figure*}

We also qualitatively compare our method with a publicly available version\footnote{\url{https://github.com/theRealSuperMario/unsupervised-disentangling/tree/reproducing_baselines}} of \cite{lorenz2019unsupervised} on LSUN Bedroom, using the standard parameters that work well on other datasets. As shown in Figure~\ref{fig:lorenz_bedroom}, their model generates trivial keypoints and fails to reconstruct the bedroom images.

\section{Editing Quality - Comparative User Study} \label{sec:supp_survey}
We designed comparisons in 4 different aspects to demonstrate our editing quality. We compare to two methods, one unsupervised keypoint-based method \cite{zhang2018unsupervised} on CelebA \cite{liu2015faceattributes}, and one mask-conditioned method \cite{zhu2020sean} on CelebA-HQ \cite{liu2015faceattributes}. For both methods, we follow their experiment design in their papers to make a fair comparison. For each question, we ask the participants to chose from 3 options: 1) ours is better; 2) theirs is better; 3) both methods have equal quality. The order of the pair to compare (ours in the first or the second) in the questions is randomly generated by the function \texttt{numpy.random.choice} of the package \texttt{Numpy} in \texttt{Python} with random seed 1. The results are illustrated in Table~\ref{tab:supp_survey}. The entire study is in anonymous form in the supplemental document at survey/index.html.

\paragraph{Study details.} We invited 23 participants answering 35 questions in total, with 5 for each of the categories above and an additional 5 comparing to randomly selected real images from the CelebA training set. The generated images are selected as follows. We generate 32 images with LatentKeypointGAN. For the baselines we take the corresponding images from their paper (as implementations are not available). We then find our image (out of the 32) that best match the baseline images (in gender, pose, head size). If multiple images are similar, we use the first one. 

\paragraph{Editing image quality.} We edited the images by swapping the appearance embedding between different images. In each question, we show one image of ours and one image of theirs. We ask the participants to compare the resulting image quality. 

\paragraph{Part disentanglement.} To compare with \cite{zhang2018unsupervised}, we moved part of the keypoints, as they did in their paper. To compare with \cite{zhu2020sean}, we exchange the part embedding between different images, as they did in their paper. In each question, we show one pair of images of ours and one pair of images of theirs. We ask the participants to choose the one with better spatial disentanglement regardless of the image quality. 

\paragraph{Identity preservation while changing expression.}
We compare identity preservation with \cite{zhang2018unsupervised}. Following their paper, we change part (or all) of the keypoints to change the expression of the face. In each question, we show the two pairs of images. Each pair contains two images, one original image, and one edited image. We ask the participants to choose the pair that can better preserve the identity of the face regardless of the image quality, as quality is assessed separately. 

\paragraph{Shape preservation while changing appearance.} We compare the shape preservation with \cite{zhu2020sean}. We edited the images by swapping the appearance embedding between different images. In each question, we show the two triplets of images. Each triplet contains three images, one shape source image and one appearance image, and one combined image. We ask the participants to choose the triplet where the combined image has the more similar shape as the shape source image regardless of the image quality.

\begin{table*}
\begin{center}
\resizebox{0.98\linewidth}{!}{%
\begin{tabular}{|l|c|c|c|c|}
\hline
Aspect & Method to compare & In favour of ours & In favour of others & Equal quality\\  \hline
Editing image quality & \cite{zhang2018unsupervised} & \textbf{92.17}\% & 0.87\% & 6.96\%\\ 
Editing image quality & SEAN \cite{zhu2020sean} & \textbf{94.78}\% & 2.61\% & 2.61\%\\ 
Part disentanglement & \cite{zhang2018unsupervised} & \textbf{67.83}\% & 5.22\% & 26.95\%\\ 
Part disentanglement & SEAN \cite{zhu2020sean} & 28.69\% & 21.74\% & \textbf{49.57}\%\\ 
Identity preservation & \cite{zhang2018unsupervised} & \textbf{55.65}\% & 14.78\% & 29.57\%\\ 
Shape preservation & SEAN \cite{zhu2020sean} & 33.91\% & \textbf{46.96}\% & 19.13\%\\ 
\hline
\end{tabular}}
\end{center}
\caption{\textbf{Survey results}. We compare 4 different aspects with other methods. The first one is the editing image quality. The second one is part disentanglement. The third one is identity preservation while changing expression. The last one is shape preservation while changing appearance.}
\label{tab:supp_survey}
\end{table*}

\paragraph{Interpretation - comparative.} This study confirms the findings of the main paper. Our method outperforms Zhang et al. in all metrics in Table~\ref{tab:supp_survey}. We also outperform SEAN in image editing quality. This confirms our claims of superior editing capability but may be surprising on the first glance since they attain a higher quality (better FID score) on unedited images. However, it can be explained with the limited editing capabilities of the mask-based approaches discussed next.

Participants give SEAN a higher shape preservation quality (47\% in favour and 19\% equal), which is expected since it conditions on pixel-accurate segmentation masks that explicitly encode the feature outline. However, the masks have the drawback that they dictate the part outline strictly, which leads to inconsistencies when exchanging appearance features across images. For instance, the strain direction of the hair and their outline must be correlated. This explains why our mask-free method attaining significantly higher image quality after editing operations (95\% preferred ours). Hence, the preferred method depends on the use case. E.g., for the fine editing of outlines SEAN would be preferred while ours is better at combining appearances from different faces. 

An additional strong outcome is that our unsupervised approach has equal disentanglement scores compared to SEAN; 50\% judge them equal, with 29\% giving preference to ours and only 22\% giving preference to SEAN. Validating that LatentKeypointGAN enables localized editing.

\paragraph{Interpretation - realism.} When comparing our GAN (without modifying keypoint location or appearance) to real images at resolution $128\times128$ of the training set, 42\% rate them as equal. Surprisingly 33\% even prefer ours over the ground truth. This preference may be because the ground truth images have artifacts in the background due to the forced alignment that are smoothed out in any of the generated ones. Overall, these scores validate that the generated images come close to real images, even though minor artifacts remain at high resolutions.

\section{Additional Implementation Details} \label{supp:archi_details}

\begin{figure}[h]
\begin{center}
   \includegraphics[width=0.98\linewidth]{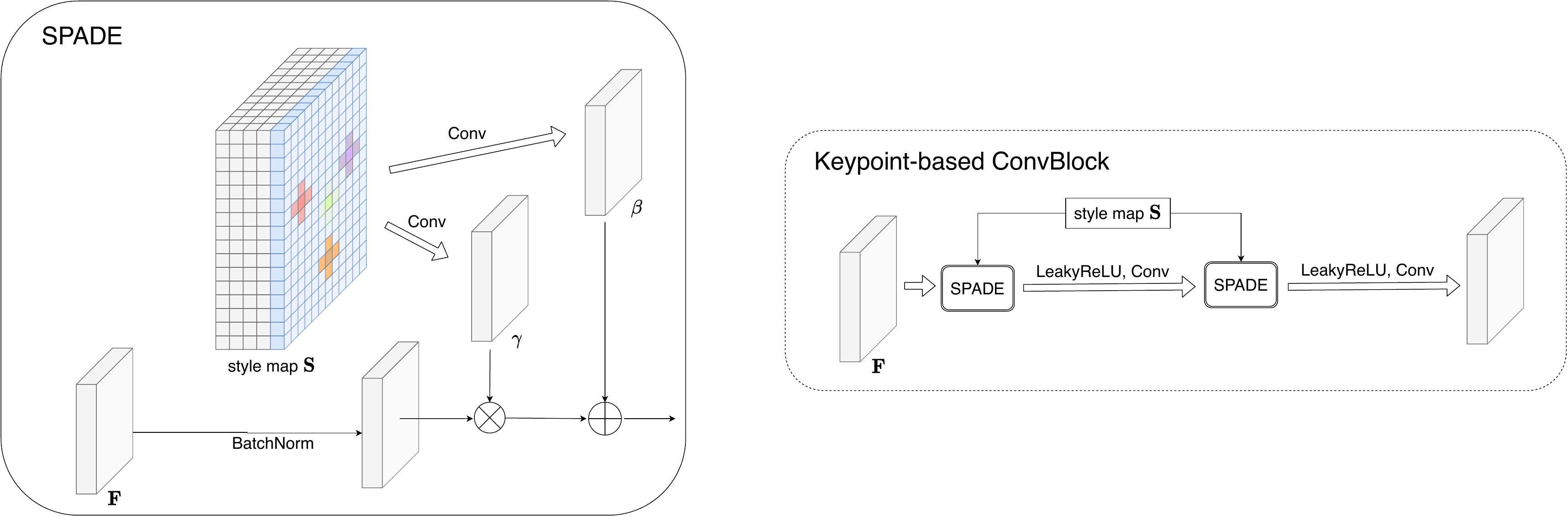}
\end{center}
   \caption{(Left) \textbf{Spatially Adaptive Normalization.} SPADE takes two inputs, feature map and style map. It first uses batch normalization to normalize the feature map. Then it uses the style map to calculate the new mean map and new standard deviation map for the feature map. (Right) \textbf{Keypoint-based ConvBlock.} }
\label{fig:spade}
\end{figure}

\paragraph{SPADE} \cite{park2019semantic} As shown in Figure~\ref{fig:spade}, SPADE takes two inputs, feature map and style map, and use the style map to calculate the mean and standard deviation, which is used to denormalize the batch normalized feature map. Formally speaking, let $\mF^i\in\myR^{N\times C_i\times H_i\times W_i}$ be a $i$-th feature map in the network for a batch of $N$ samples, where $C_i$ is the number of channels. Here we slightly abuse the notation to denote $N$ batched style maps of size $(H_i,W_i)$ as $\mS^i\in\myR^{N\times (K+1)D_\text{embed}\times H_i\times W_i}$. The same equation as for BatchNorm \cite{ioffe2015batch} is used to normalize the feature map, but now the denormalization coefficients stem from the conditional map, which in our case is the processed style map. Specifically, the resulting value of the spatial adaptive normalization is
\begin{equation}
A^i_{n,c,y,x}(\mS,\mF) =    \gamma^i_{c,y,x}(\mS^i_n)\frac{\mF^i_{n,c,y,x}-\mu^i_c}{\sigma^i_c}+\beta^i_{c,y,x}(\mS^i_n),
\end{equation}
where $n\in\{1,...,N\}$ is the index of the sample, $c\in\{1,...,C\}$ is the index of channels of the feature map, and $(y,x)$ is the pixel index. 
The $\mu^i_c$ and $\sigma^i_c$ are the mean and standard deviation of channel $c$.
% \comment{,
% \begin{equation}
% \begin{aligned}
%     \mu^i_c&=\frac{1}{NH_iW_i}\sum_{n,y,x}\mF^i_{n,c,y,x}\\
%     \sigma^i_c&=\sqrt{\frac{1}{NH_iW_i}\sum_{n,y,x}\bigg(\mF^i_{n,c,y,x}-\mu^i_c\bigg)^2},
% \end{aligned}
% \end{equation}
% where $N$ is the batch size.}
%
The $\gamma^i_{c,y,x}(\mS^i_n)$ and $\beta^i_{c,y,x}(\mS^i_n)$ are the parameters to denormalize the feature map. They are obtained by applying two convolution layers on the style map $\mS^i_n$.

\paragraph{Initialization.} We use Kaiming initialization \cite{he2015delving} for our network. We initialize the weights of the last layer of the MLP that generates the keypoint coordinates to 0.05x the Kaiming initialization.

\begin{figure*}[t]
\begin{center}
   \includegraphics[width=0.98\linewidth]{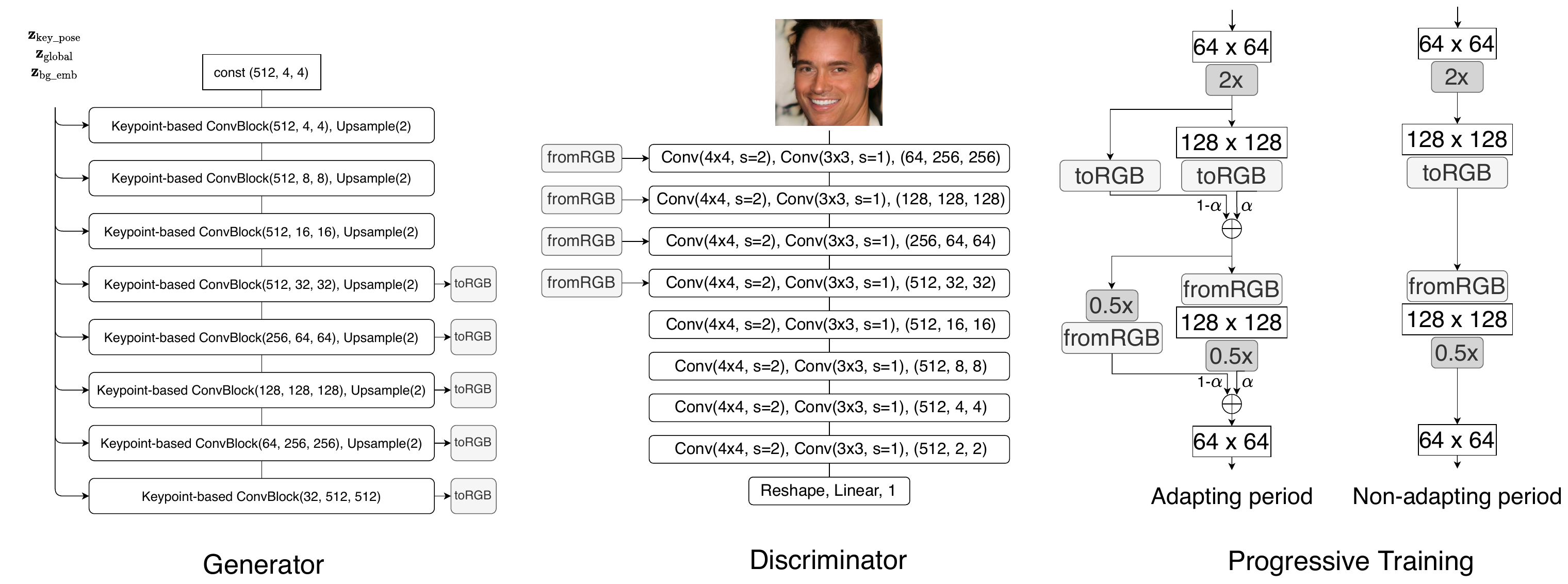}
\end{center}
   \caption{\textbf{Detailed architecture.} (Left) \textbf{LatentKeypointGAN generator.} The numbers in the parenthesis is the output dimension of the Keypoint-based ConvBlock. For example, (512, 4, 4) means the output feature map has a resolution of $4\times 4$ and the channel size is 512. The toRGB blocks are $1\times1$ convolutions to generate the RGB images with the same resolution as corresponding feature maps. (Middle) \textbf{LatentKeypointGAN discriminator.} The number in the last parenthesis is the output dimension. For example, (512, 4, 4) means the output feature map has a resolution of $4\times 4$ and the channel size is 512. At each resolution, we apply two convolutions, one with stride 2 to downsample feature maps and one with stride 1 to extract features. Leaky ReLU \cite{maas2013rectifier} is used after all convolutions except the linear layer in the last.. (Right) \textbf{Progressive Training.}. The adapting period is the same as PGGAN \cite{karras2018progressive} and StyleGAN \cite{karras2019style}. In the non-adapting period, we do not use the linear combination.}
\label{fig:detail_archi}
\end{figure*}

\paragraph{Progressive Growing Training} We adopt progressive growing training \cite{karras2018progressive} to allow larger batch size, which helps both on keypoint localization and local appearance learning. This is likely due to the BatchNorm that is essential in SPADE. We also tried to replace the BatchNorm in SPADE with LayerNorm \cite{ba2016layer} and PixelNorm \cite{karras2019style}, but both of them cause mode collapse. We use a scheduled learning rate for the Keypoint Generator $\cK$. As illustrated in Figure~\ref{fig:detail_archi}, at each resolution stage, the training is divided into adapting period and non-adapting period. We set the learning rate of $\cK$ to zero in the adapting period and back to normal in the non-adapting period. In the adapting period, the training follows PGGAN \cite{karras2018progressive} where the feature map is a linear combination of the larger resolution RGB image and current resolution RGB image. The coefficient $\alpha$ gradually increases from 0 to 1. At the end of the adapting period, the network is fully adapted to generate higher resolution images. In the non-adapting period, the network generates high-resolution images without the linear combination. Following StyleGAN \cite{karras2019style}, we start from a $4\times 4\times 512$ learned constant matrix, which is optimized during training and fixed during testing. We use the keypoint-based ConvBlock and bilinear upsampling to obtain feature maps with increasing resolutions. Unlike PGGAN \cite{karras2018progressive} and StyleGAN \cite{karras2019style}, who generating RGB images from feature maps of all resolutions (from $4\times4$ to $1024\times 1024$), we start generating RGB images from the feature maps of at least $64\times 64$ resolution. This is possible with the keypint generator and its spatially localized embeddings taking over the role of low feature maps. It helps to locate the keypoints more accurately.

\paragraph{Generator.} We illustrate the LatentKeypointGAN generator in Figure~\ref{fig:detail_archi}. The output image is linearly combined by the output of \textit{toRGB} block, where the weights depend on the training stage.

\paragraph{Discriminator.}
We illustrate the discriminator in Figure~\ref{fig:detail_archi}. For each resolution, we use two convolutions followed by Leaky ReLU \cite{maas2013rectifier}. The first convolution has a kernel size $4\times 4$ and stride 2 to downsample the feature map to 0.5x. The second convolutions have a kernel size $3\times 3$ and stride 1 to extract features.

\paragraph{Hyperparameter Setting}
For all experiments in image generation, we use leaky ReLU \cite{maas2013rectifier} with a slope 0.2 for negative values as our activation function. We use ADAM optimizer \cite{KingmaB14} with $\beta_1=0.5$ and $\beta_2=0.9$. We set the learning rate to 0.0001 and 0.0004 for generator and discriminators, respectively \cite{heusel2017gans}. We start from generating $64\times 64$ images in the progressive training. The batch size for $64^2, 128^2, 256^2, 512^2$ images are $128,64,32,8$, respectively. We set $\lambda_\text{gp}=10$ and $\lambda_\text{bg}=100$. We set $D_\text{noise}=256$ and $D_\text{embed}=128$ for all experiments unless otherwise stated (in ablation tests).

We lists the different $\tau$s and different background setting for all experiments in Table~\ref{tab:setting_datasets}.  In CelebA-HQ and FFHQ, the foreground is naturally disentangled from the background. The face can be freely moved on the image. However, in the Bedroom dataset, all objects and their parts are strongly correlated.
For example, the bed cannot be moved to the ceiling, and the window cannot be moved to the floor. Therefore, we treat every object in the bedroom as a key part, even the floor, but the possible motion is restricted to plausible locations (see the supplementary video). A separate background embedding does not make sense. Therefore, we set the background ($\mH^{bg}=0$) and the background loss $\lambda_\text{bg}=0$ for the experiments on the Bedroom dataset.

\begin{table}
\begin{center}
% \resizebox{0.75\linewidth}{!}{%
\begin{tabular}{|l|c|c|}
\hline
Dataset & background module and loss & $\tau$\\ 
\hline
CelebA-HQ & yes & 0.01\\ 
FFHQ & yes & 0.01\\ 
Bedroom & no & 0.01\\
BBC Pose & yes & 0.025\\  \hline
\end{tabular}
\end{center}
\caption{\textbf{Setting for different datasets}. For the Bedroom dataset, we do not use the background module and loss. For the BBC Pose dataset, we use $\tau=0.025$.}
\label{tab:setting_datasets}
\end{table}
\section{Additional Ablation Tests} \label{sec:ablation_test}

\subsection{Ablation Test on the Neural Network Architecture}

We provide here additional insights into the ablation study summarised in the main paper.
%We show the detection result on MAFL and FID score on FFHQ in Table~\ref{tab:ablation_archi} for different architectures we tested below. The detection follows Section~\ref{sec:unsupervised_keypoints_discovery}. The FID is calculated at resolution $256\times 256$, by generated 50k images and the resized original dataset. Note that it is different from the main paper for simplicity since the goal of this part is ablation tests instead of comparing with others. 

\begin{figure*}[t]
\begin{center}
  \includegraphics[width=0.98\linewidth]{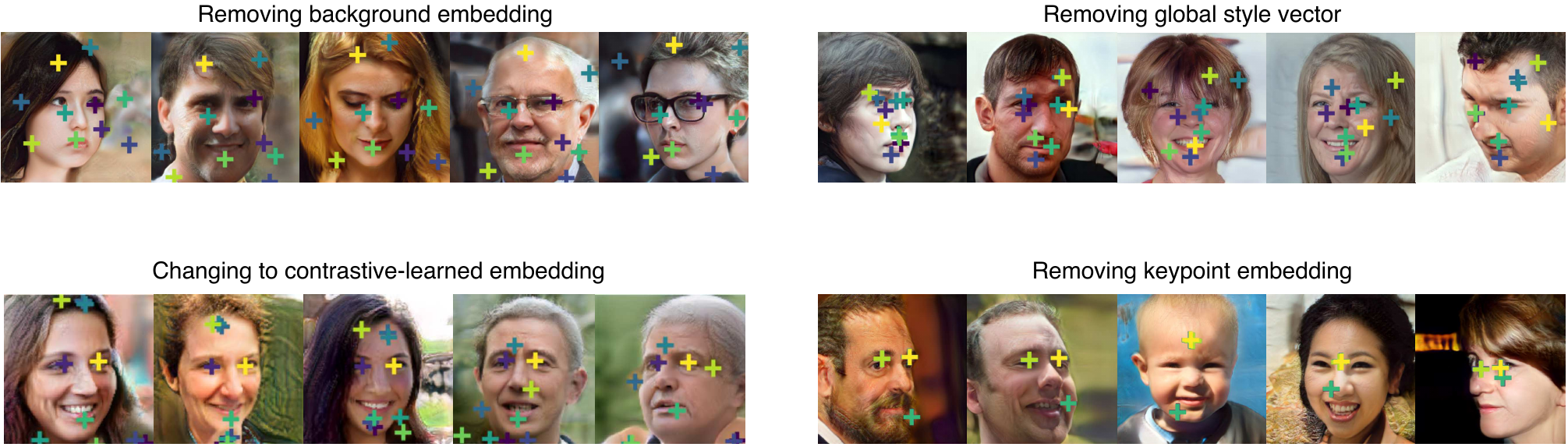}
\end{center}
  \caption{\textbf{Ablation study on architecture}. We show the keypoints for different architectures.}
\label{fig:supp_ablation_archi}
\end{figure*}

\paragraph{Removing background embedding.} We remove the background embeddings from our architecture ($z_\text{bg\_emb}$ and $\vw_\text{bg}$). In this case, the keypoint embedding controls the whole appearance of the image. In addition, as shown in Figure~\ref{fig:supp_ablation_archi}, the keypoints are not exactly located at where a human would place them, though they are still consistent among different images and views.

\paragraph{Removing global style vector.} \label{sec:supp_removing_global} We remove the global style vector $\vw_\text{global}$. Therefore, all the keypoint embeddings are constant. Only keypoint location and background embedding are different among the images. In this case, the keypoint embedding works equivalent to one-hot encoding, and cannot fully capture the variations on the key parts. Therefore, it leads to inaccurate keypoints, as shown in Figure~\ref{fig:supp_ablation_archi}. Furthermore, we observed that without $\vw_\text{global}$, the network hides the appearance information in the keypoint location, leading to unwanted entanglement of pose and appearance.
% We believe this is because that the global style vector $\vw_\text{global}$ not only controls the appearance but also works as a noise. Without this noise, the network treats keypoints as anchors to plot the images. This explains why some keypoints are not consistent with different appearances. This can also harm detection accuracy and image quality.

\paragraph{Changing keypoint embedding generation.} We change the keypoint embedding generation in two ways. The first way is generating constant embedding $\vw_\text{const}^j$ and global style vector $\vw_\text{global}$ just as before and then add them elementwisely instead of multiplying them. Formally speaking, for each keypoint $j$, its corresponding embedding is
\begin{equation}
    \vw^j=\vw_\text{global} \oplus \vw^j_\text{const},
    \label{eq:noise combination add}
\end{equation}
where $\oplus$ means elementwise addition. This gives slightly higher detection accuracy but lower image quality. We observe that in this case, the background controls the foreground appearance. However, different from Removing global style vector in Section~\ref{sec:supp_removing_global}, the appearance information is not hidden in keypoint locations. We believe this is because that $\vw_\text{global}$ works as noise to avoid the network from hiding foreground appearance information in keypoint location. As a result of good disentanglement of appearance and keypoint location, the keypoint detection accuracy slightly increases. However, again, in this setting, the keypoint embedding cannot fully capture the variations of the key parts. Therefore, the background takes the control of appearance and we discarded this avenue.

\begin{figure*}[t]
\begin{center}
   \includegraphics[width=0.98\linewidth]{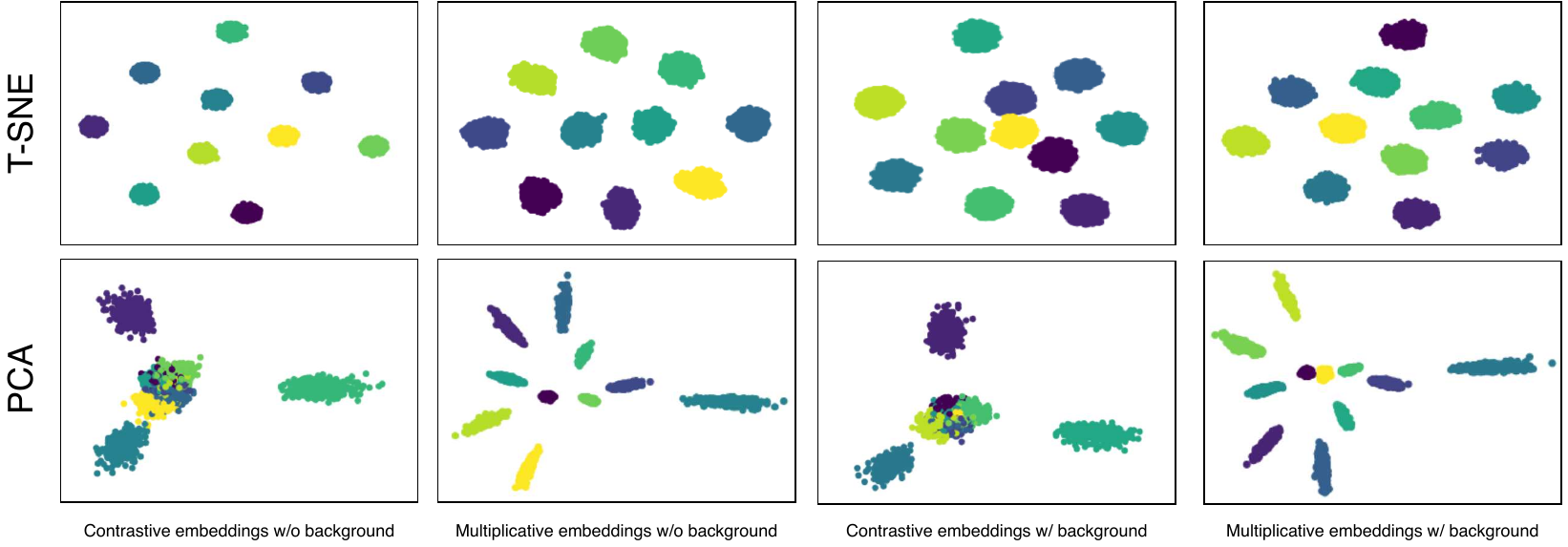}
\end{center}
   \caption{\textbf{Ablation study on multiplicative embedding}. We show the T-SNE and PCA visualization of embeddings learned on FFHQ. The first two column shows keypoint embeddings and the last two column shows keypoint embeddings and background embedding.}
\label{fig:supp_ablation_embedding_way}
\end{figure*}

\label{supp:contrastive_learning}
The second way is to generate $[\vw^j]_{j=1}^K$ together from $\vz_\text{kp\_app}$ using a single MLP. In this case, there is no constant embeddings or global style vector. To force the embedding of the same keypoint to be similar, and the embedding for different keypoints to be different, we tried Supervised Contrastive Losses \cite{khosla2020supervised},
\begin{equation}
\begin{aligned}
    &\mathcal L_\text{contrastive}(\mathcal G) = \\
    &-\sum_{j\in J}\frac{1}{|K(j)|}\sum_{k\in K(j)}\log\frac{\exp(\vw^j \cdot \vw^k / T)}{\sum_{a\in A(j)} \exp(\vw^j \cdot \vw^a / T)},
\end{aligned}
    \label{eq:noise combination contrastive}
\end{equation}
where 
$$
\begin{aligned}
    A(j)&=\{i: \vw^i, \vw^j \text{are in the same batch}\}&\\
    K(j)&=\{i: \vw^i, \vw^j \text{belong to the same keypoint}&\\
    &                                \text{in the same batch}\}&\\
    J&=\{\text{indices of all } \text{keypoint embeddings}&\\
    &\text{in the same batch}\}&
\end{aligned}
$$
As shown in Figure~\ref{fig:supp_ablation_archi}, the keypoints are neither on the key parts nor consistent. We further visualize the embeddings with T-SNE and PCA in Figure~\ref{fig:supp_ablation_embedding_way}. Although the contrastive learned embedding has comparable T-SNE with our multiplicative design, the PCA shows that our multiplicative embedding is linearly separable while contrastive learned embedding is not. Hence, we demonstrate that our original design of elementwise products is simple and effective.

\paragraph{Removing keypoint embedding.} We remove the keypoint embedding $\vw^j$ entirely. In this case, we only have background embedding $\vw_\text{bg}$ and the keypoint location. Thus, instead of generating the style map $\mS$, we directly concatenate the keypoint heatmaps $\{\mH^k\}_{k=1}^K$ and the broadcasted background style map to generate the style map without keypoint embedding. As shown in Figure~\ref{fig:supp_ablation_archi}, the keypoints are not meaningful or consistent. The keypoint location hides the appearance and is entangled with the background severely.

\paragraph{Removing keypoints.} If we remove the keypoints, then SPADE \cite{park2019semantic} degenerates to AdaIN \cite{huang2017arbitrary}. Instead of using a style map $\mS$ (2D), we now use a style vector (1D), which is the background embedding. In this case, we do not have the ability to control the generated images locally.

% \begin{table}
% \begin{center}
% % \resizebox{0.85\linewidth}{!}{%
% \begin{tabular}{|l|c|c|}
% \hline
% Method & Keypoint detection error on MAFL $\downarrow$ & FID score on FFHQ $\downarrow$\\ 
% \hline
% full model & 5.85\% & 11.17\\ 
% w/o background & 6.43\% & 14.45\\ 
% w/o global style vector & 6.76\% & 13.61\\ 
% adding global style vector & 5.29\% & 16.83\\ 
% contrastive keypoint embedding & 7.53\% & 18.29 \\
% w/o keypoint embedding & 22.81\% & 12.89\\ 
% w/o keypoint & - & 15.73\\ 
% \hline
% \end{tabular}
% \end{center}
% \caption{\textbf{Quantitative ablation test on architecture.} For both metrics, the lower means better.}
% \label{tab:supp_ablation_archi}
% \end{table}

\subsection{Ablation Tests on the Hyperparameters}

\paragraph{Ablation test on the dimension of embeddings.} Different numbers of embedding dimensions make the expressive power vary. As shown in Table~\ref{tab:ablation_n_embedding}, larger $D_\text{embed}$ leads to larger error on MAFL but lower (better) FID. We use $D_\text{embed}=128$ in our main paper because the increase in error is small but the decrease of FID is significant.

\begin{figure*}[t]
\begin{center}
   \includegraphics[width=0.98\linewidth]{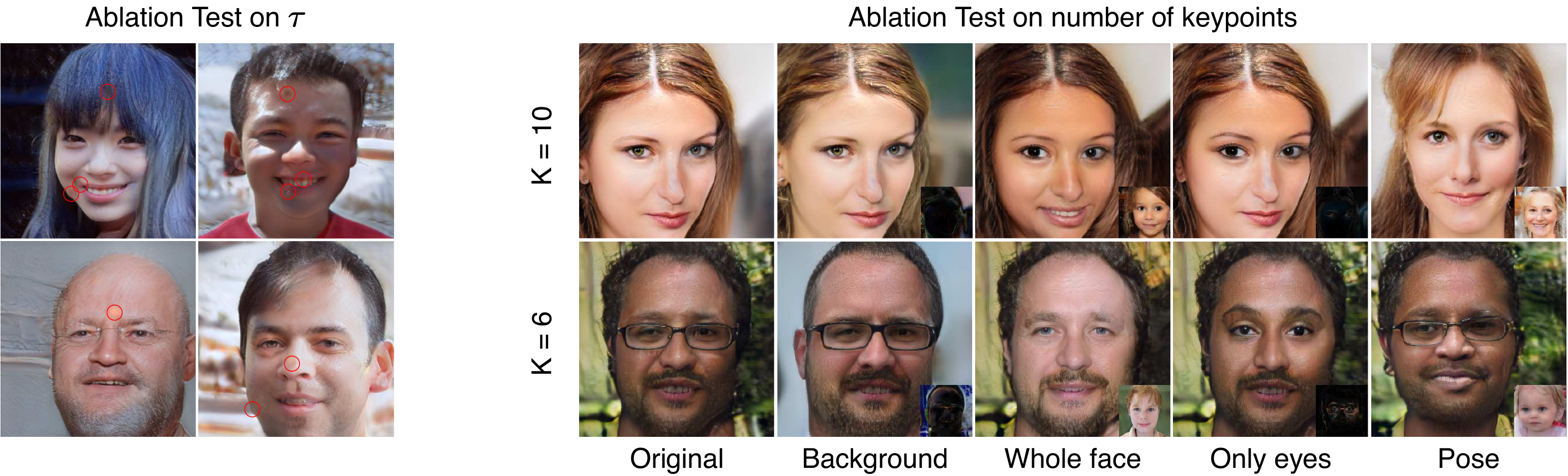}
\end{center}
   \caption{\textbf{Ablation study on hyperparameters}. (Left) Face generation with on FFHQ with $\tau=0.002$. We use the red circle to mark the artifacts in the images. (Right) Face generation on FFHQ with number of keypoints 10 (top) and 6 (bottom). More keypoints lead to a stronger influence of the keypoint embedding. However, the 6-keypoint version still provides control, e.g., glasses, nose type, and pose. From left to right: original image, replaced background (difference map overlayed), replaced keypoint embeddings (target image overlayed), exchanged eye embeddings, and keypoint position exchanged.}
\label{fig:ablation_hyperparameters}
\end{figure*}

\paragraph{Ablation Test on $\tau$}
A too-small value for $\tau$ does not influence the image and will cause artifacts as shown in Figure~\ref{fig:ablation_hyperparameters}. A too-large value for $\tau$ will disable the background embedding and control the background.

\begin{table*}
\begin{center}
\resizebox{0.98\linewidth}{!}{%
\begin{tabular}{|c|c|c|}
\hline
Dimension of embeddings $D_\text{embed}$ & Keypoint detection error on MAFL $\downarrow$ & FID score on FFHQ $\downarrow$\\ 
\hline
16 & 4.61\% & 28.85\\ 
32 & 4.92\% & 26.14\\ 
64 & 5.66\% & 27.64\\ 
128 & 5.85\% & 23.50\\ 
\hline
\end{tabular}}
\end{center}
\caption{\textbf{Quantitative ablation test on dimension of embeddings.} For both metrics, the lower means better.}
\label{tab:ablation_n_embedding}
\end{table*}

\paragraph{Ablation test on the number of keypoints.} By selecting different numbers of keypoints, we can achieve different levels of control. In the second row of Figure~\ref{fig:ablation_hyperparameters}, we use 6 keypoints rather than the default 10. Thereby, keypoints have a smaller area of effect. We observe that the background encoding then takes a larger role and contains the encoding of hair and beard, while the keypoints focus only on the main facial features (nose, mouth, and eyes).

\begin{figure*}[t]
\begin{center}
   \includegraphics[width=0.98\linewidth]{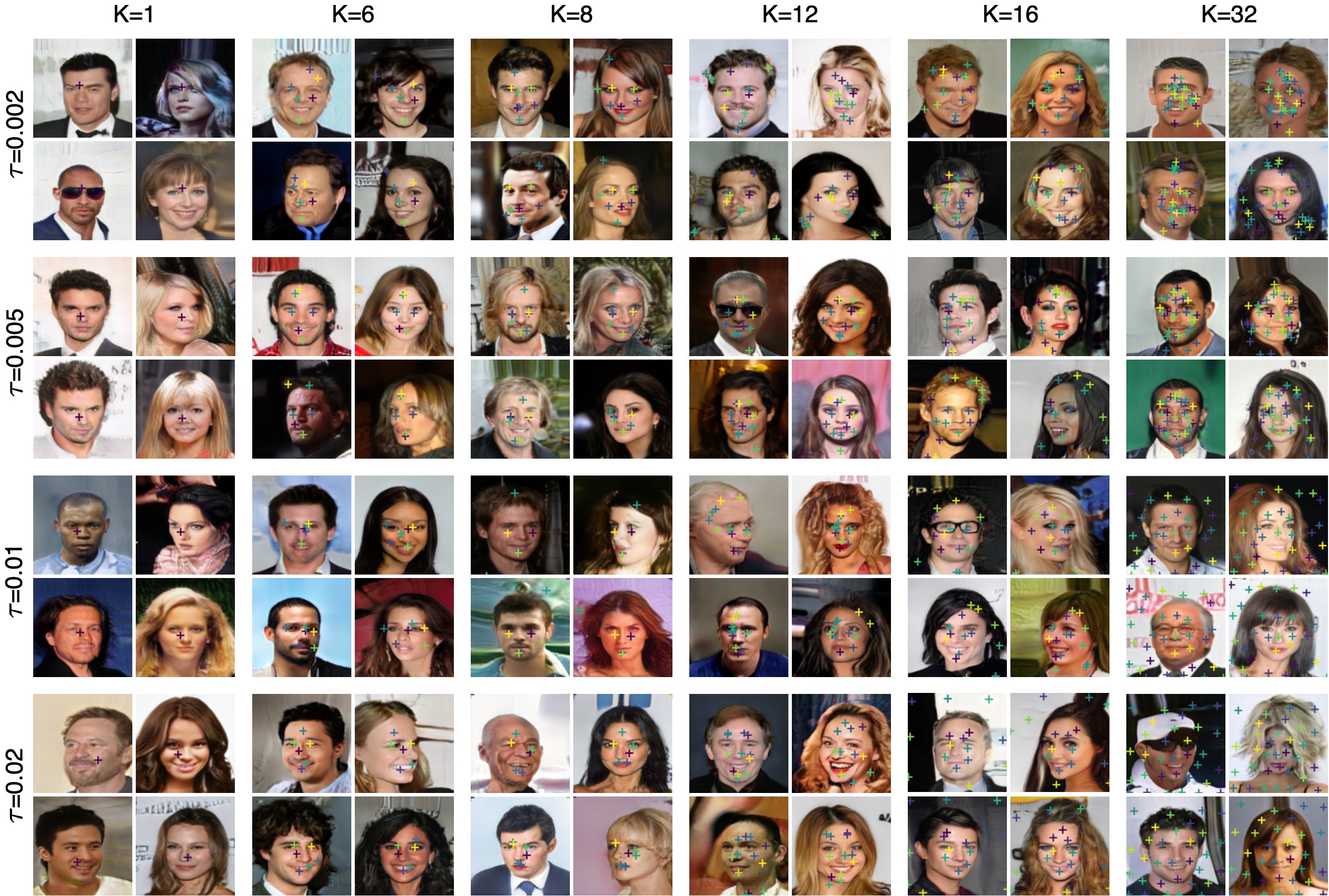}
\end{center}
   \caption{\textbf{Visualization for different combinations of number of keypoints and keypoint size $\tau$}.  If both, the number of keypoints and the keypoint size $\tau$, are small (top left), the keypoint is trivial. If both of them are large (bottom right), the keypoints distribute uniformly over the images instead of focusing on parts.}
\label{fig:visualization_combination_n_keypoints_tau}
\end{figure*}

\begin{figure*}[t]
\begin{center}
   \includegraphics[width=0.98\linewidth]{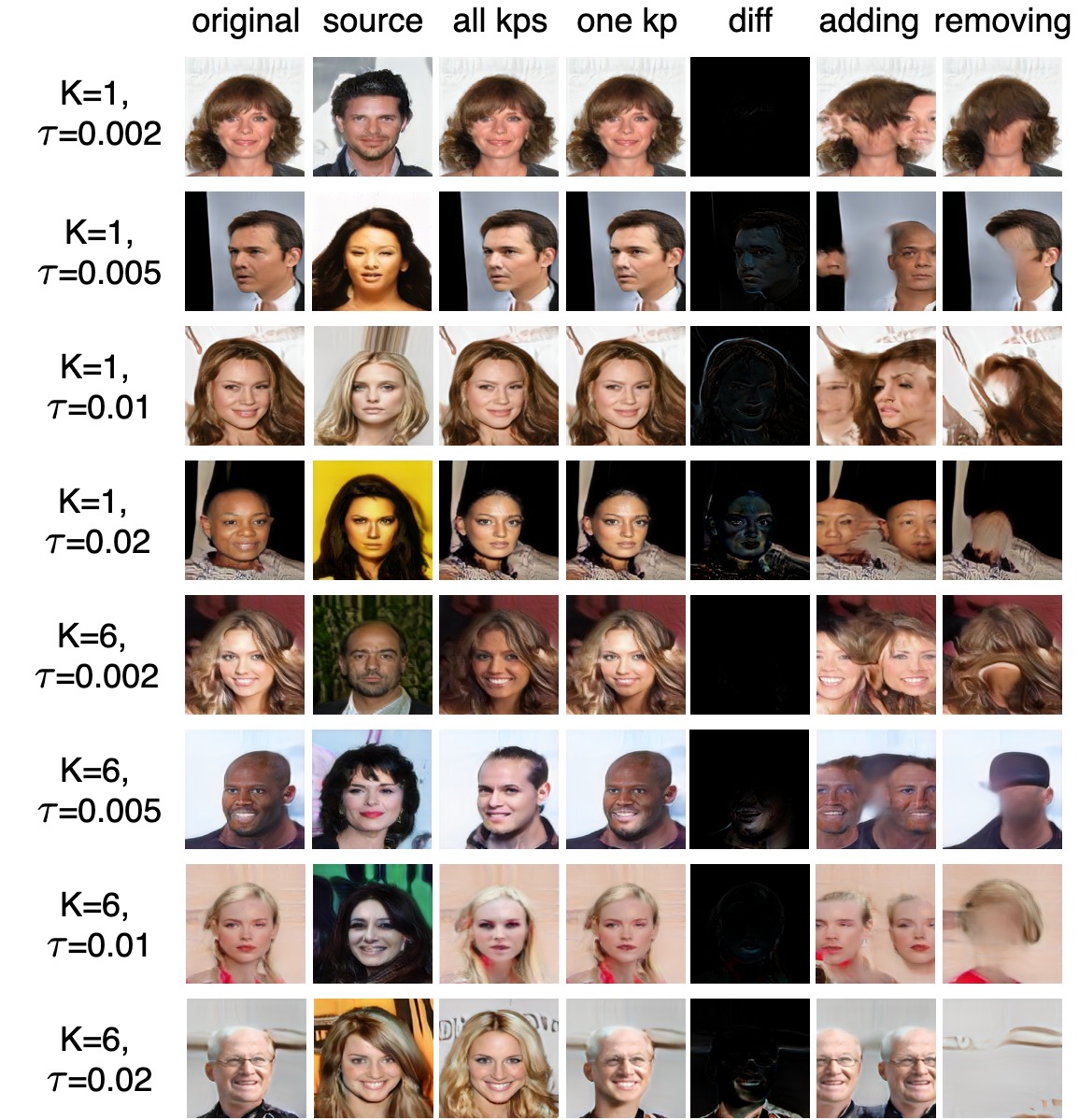}
\end{center}
   \caption{\textbf{Editing on different combinations of number of keypoints $K$ and keypoint size $\tau$}. K=1, 6. \textbf{Column 1}: original image; \textbf{column 2}: part appearance source image used to swap appearance; \textbf{column 3}: the combined image with shape from the original images and the appearance from the part appearance source image; \textbf{column4}: we randomly 
  swap a single keypoint close to the mouth; \textbf{column 5}: resulting difference map when changing the keypoint in the 4th column; \textbf{column 6}: move the face to the left and add another set of keypoints on the right; \textbf{column 7}: removing all keypoints. If $\tau=0.0002$, the keypoints are trivial, and cannot be used to change appearance. When $K=1$, the keypoint also only have limited control even if $\tau=0.02$. The combination of $K=6, \tau=0.005$ gives good spatial disentanglement.}
\label{fig:editing_combination_n_keypoints_tau}
\end{figure*}

\begin{figure*}[t]
\begin{center}
   \includegraphics[width=0.98\linewidth]{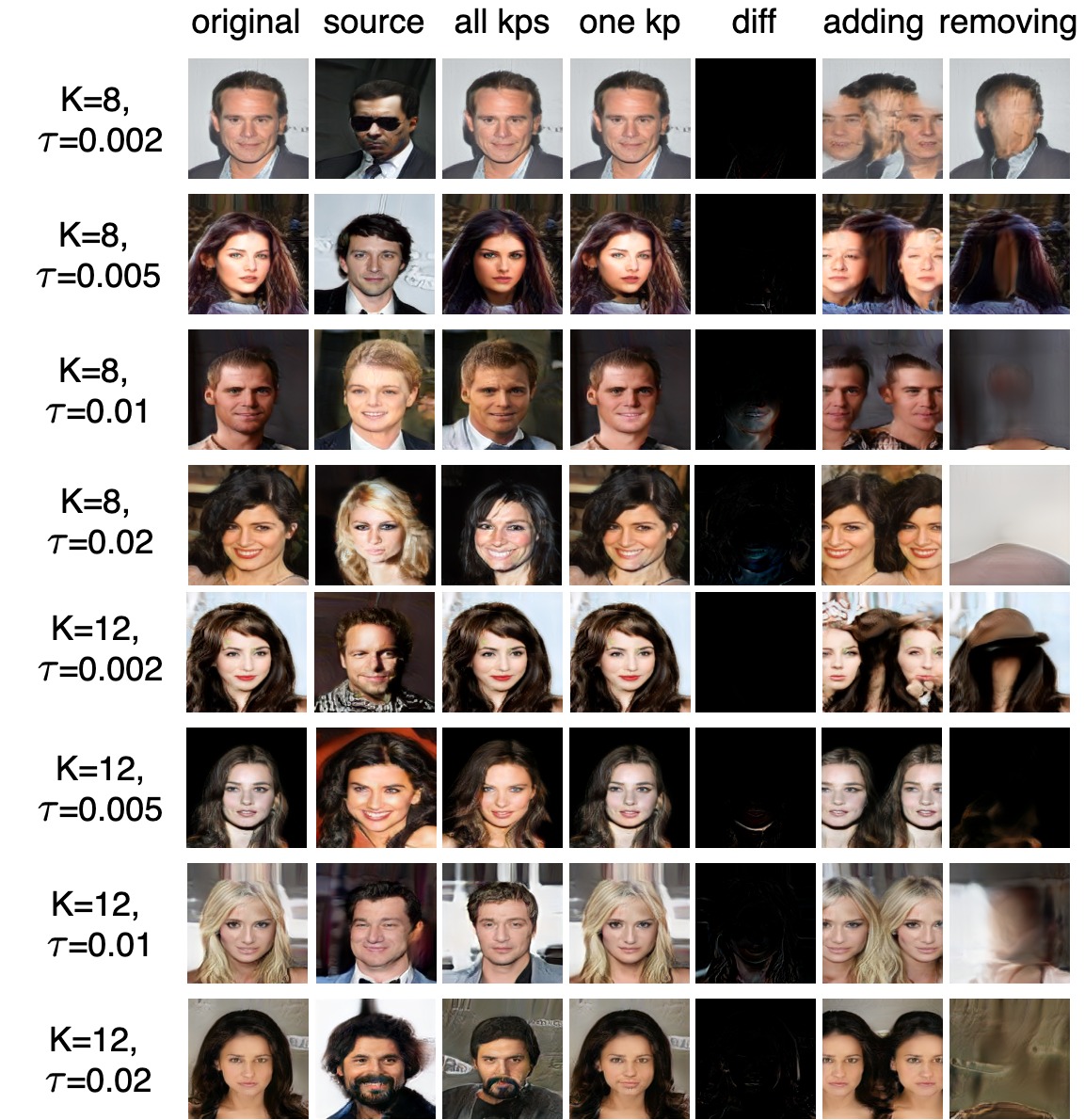}
\end{center}
   \caption{\textbf{Editing on different combinations of number of keypoints $K$ and keypoint size $\tau$}. K=8,12. For a small $\tau=0.0002$, the keypoint is trivial. When $\tau$ is large the background is entangled ($K=8, \tau=0.02$) in some cases. We found the combinations of ($K=12, \tau=0.0005$) and ($K=12,\tau=0.01$) both give the best editing controllability.}
\label{fig:editing_combination_n_keypoints_tau2}
\end{figure*}

\paragraph{Ablation test on the combination of number of keypoints and $\tau$.} The impact of keypoints depends on the combination of number of keypoints and $\tau$. We test the pairwise combination between $K=1,6,8,12,16,32$ and $\tau=0.002, 0.005, 0.01, 0.02$. The FID is listed in Table~\ref{tab:ablation_n_keypoints_tau_fid} and the detection error is listed in Table~\ref{tab:ablation_n_keypoints_tau_acc}. The image quality does not change much for different combinations. We illustrate samples of keypoints of each combination in Figure~\ref{fig:visualization_combination_n_keypoints_tau} and editing results in Figure~\ref{fig:editing_combination_n_keypoints_tau} and Figure~\ref{fig:editing_combination_n_keypoints_tau2}. If both the number of keypoints and $\tau$ are small, e.g., $K=1$, and the $\tau=0.002$, then the background controls both foreground appearance and pose, and the keypoints are trivial. If both, the number of keypoints and $\tau$, are large, e.g., $K=32$, and $\tau=0.02$, then the keypoint appearance controls background and pose. While the model degenerates in extreme cases, we found the model to be robust for a wide range of values, i.e., $K=8, 12, 16$ and $\tau=0.005, 0.01$. We summarize all the cases in Table~\ref{tab:ablation_n_keypoints_tau}.

\begin{table*}[t]
\begin{center}
\begin{tabular}{|c|c|c|c|c|c|c|}
\hline
  & $K=1$ & $K=6$ & $K=8$ & $K=12$ & $K=16$ & $K=32$\\ 
\hline
$\tau=0.002$ & 0.07 & 0.20 & 0.19 & 0.19 & 0.22 & 0.32\\ 
$\tau=0.005$ & 0.26 & 0.35 & 0.54 & 0.42 & 0.38 & 0.47\\ 
$\tau=0.01$ & 0.22 & 0.30 & 0.48 & 0.52 & 0.50 & 0.38\\ 
$\tau=0.02$ & 0.20 & 0.48 & 0.48 & 0.42 & 0.39 &  0.37\\ 
\hline
\end{tabular}
\end{center}
\caption{\textbf{CPD scores of ablation tests on number of keypoints $K$ and keypoint size $\tau$ on CelebA of resolution $128\times 128$.} The higher means better. 
% Neither $K$ or $\tau$ significantly influence the image quality. Interestingly, the small artifacts when $\tau=0.002$ in Figure~\ref{fig:ablation_hyperparameters} does not neither significantly influence the image quality.
}
\label{tab:ablation_n_keypoints_tau_cpd}
\end{table*}

\begin{table*}[t]
\begin{center}
\begin{tabular}{|c|c|c|c|c|c|c|}
\hline
  & $K=1$ & $K=6$ & $K=8$ & $K=12$ & $K=16$ & $K=32$\\ 
\hline
$\tau=0.002$ & 19.35 & 18.94 & 17.77 & 17.69 & 20.44 & 19.29\\ 
$\tau=0.005$ & 18.39 & 18.28 & 18.42 & 18.38 & 20.31 & 18.72\\ 
$\tau=0.01$ & 19.49 & 19.60 & 20.31 & 18.14 & 19.25 & 17.91\\ 
$\tau=0.02$ & 19.11 & 18.80 & 20.28 & 19.34 & 19.58 & 18.17\\ 
\hline
\end{tabular}
\end{center}
\caption{\textbf{FID scores of ablation tests on number of keypoints $K$ and keypoint size $\tau$ on CelebA of resolution $128\times 128$.} The lower means better. Neither $K$ or $\tau$ significantly influence the image quality. Interestingly, the small artifacts when $\tau=0.002$ in Figure~\ref{fig:ablation_hyperparameters} does not neither significantly influence the image quality.}
\label{tab:ablation_n_keypoints_tau_fid}
\end{table*}

\begin{table*}[t]
\begin{center}
\begin{tabular}{|c|c|c|c|c|c|c|}
\hline
  & $K=1$ & $K=6$ & $K=8$ & $K=12$ & $K=16$ & $K=32$\\ 
\hline
$\tau=0.002$ & 11.59\% & 7.86\% & 6.78\% & 7.09\% & 6.00\% & 5.35\%\\ 
$\tau=0.005$ & 8.65\% & 7.28\% & 6.39\% & 5.24\% & 5.13\% & 4.11\%\\ 
$\tau=0.01$ & 8.43\% & 7.91\% & 7.97\% & 6.06\% & 7.37\% & 8.84\%\\ 
$\tau=0.02$ & 8.71\% & 6.26\% & 7.13\% & 5.16\% & 6.80\% & 8.26\%\\ 
\hline
\end{tabular}
\end{center}
\caption{\textbf{Normalized Error of ablation tests on number of keypoints $K$ and keypoint size $\tau$ on CelebA of resolution $128\times 128$.} For $\tau=0.002, 0.005$, the error decreases as $K$ increases while for $\tau=0.01, 0.02$, the error first decreases and then increases. If both of them are large, e.g., $K>16, \tau>0.01$, the appearance is entangled with the keypoints which results in a larger error.}
\label{tab:ablation_n_keypoints_tau_acc}
\end{table*}

\begin{table*}[t]
\begin{center}
\begin{tabular}{|c|c|c|c|c|c|c|}
\hline
  & $K=1$ & $K=6$ & $K=8$ & $K=12$ & $K=16$ & $K=32$\\ 
\hline
$\tau=0.002$ & T & T & T & T & T & T\\ 
$\tau=0.005$ & T & \checkmark\checkmark & \checkmark\checkmark & \checkmark\checkmark & \checkmark\checkmark & \checkmark\checkmark\\ 
$\tau=0.01$ & T & \checkmark & \checkmark & \checkmark\checkmark & \checkmark\checkmark & E\\ 
$\tau=0.02$ & \checkmark & \checkmark & \checkmark & \checkmark & E & E\\ 
\hline
\end{tabular}
\end{center}
\caption{\textbf{Keypoint controllability}. T denotes trivial keypoint, i.e., the background controls the entire image. E means entangled pose, appearance and background. \checkmark\checkmark means disentangled control and \checkmark means inferior disentanglement, where one of the pairs \{(pose, appearance), (pose, background), (appearance, background) is entangled.\}. For a small keypoint size of $\tau=0.0002$ the model always gives trivial keypoints. With a large number of keypoints and a large keypoint size, i.e., $K>16$ and $\tau>0.01$, our model gives entangled representations. Our model is robust in the range of $K\in[8,16]$ and $\tau\in[0.005,0.01]$.}
\label{tab:ablation_n_keypoints_tau}
\end{table*}

\begin{figure*}[t]
\begin{center}
   \includegraphics[width=0.98\linewidth]{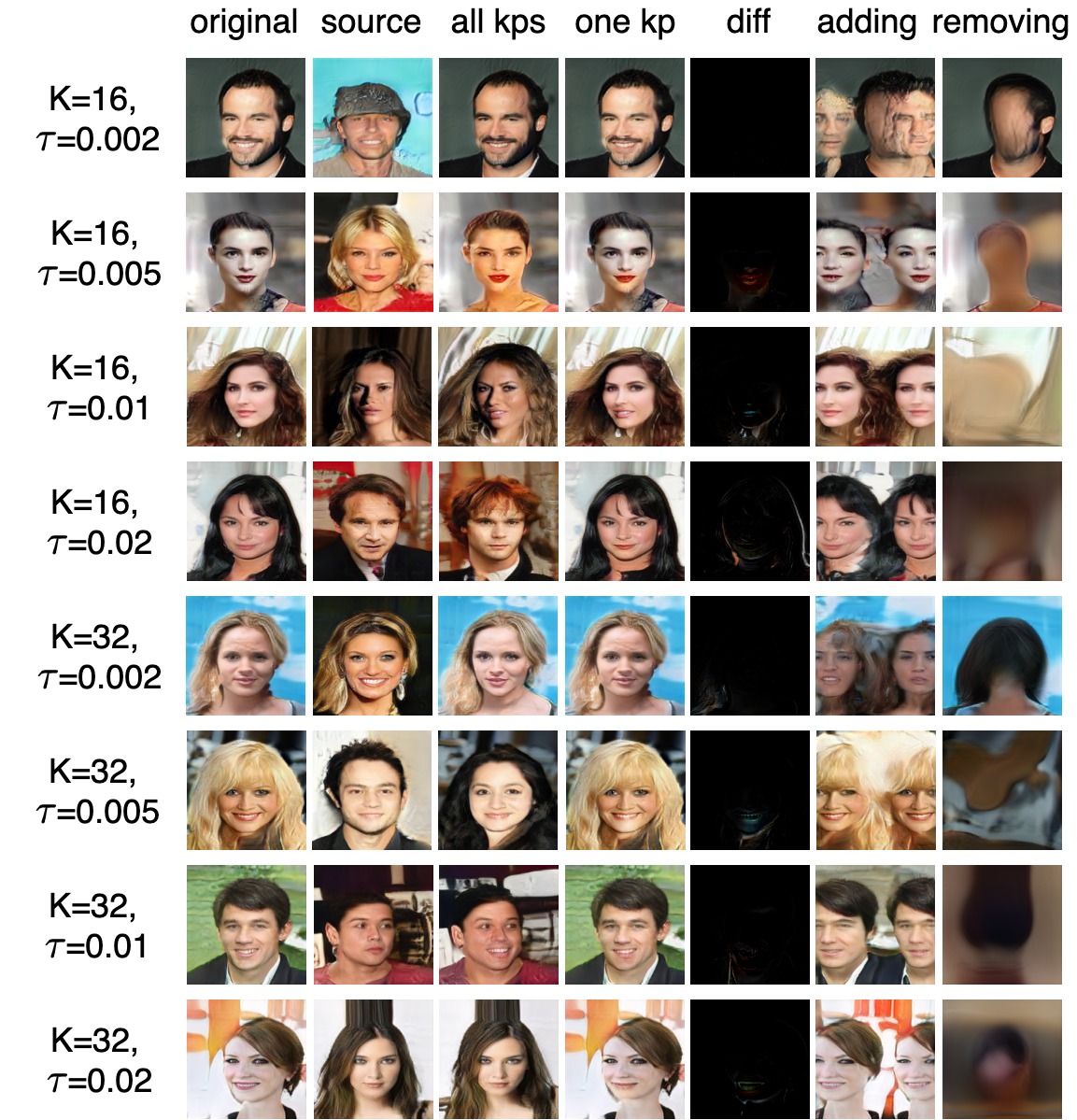}
\end{center}
   \caption{\textbf{Editing on different combination of number of keypoints $K$ and keypoint size $\tau$}. K=16,32. Extreme small $\tau$ ($\tau=0.0002$) constantly gives trivial keypoints even if $K$ is large ($K=32$). When both, $K$ and $\tau$, are large ($K=32, \tau>0.01$), the keypoint embeddings control the background and the pose. We found the combinations of ($K=16, \tau=0.0005$) and ($K=32,\tau=0.005$) both give the best editing controllability.}
\label{fig:editing_combination_n_keypoints_tau3}
\end{figure*}

\subsection{Ablation Tests on the Training Strategy}

\begin{figure}
%  \vspace{-40pt}
% \setlength{\abovecaptionskip}{-2pt}
  \includegraphics[width=1\linewidth]{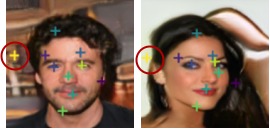}
  \caption{\textbf{Ablation Test on the Background loss.} Without it, some keypoints control background features.
%   The background loss is not essential but stabilizes the keypoints. Without it, the keypoints still work but some may move onto the background.
  }
\label{fig:ablation_bg_loss}
%  \vspace{100pt}
\end{figure}

\begin{figure*}[t]
\begin{center}
   \includegraphics[width=0.98\linewidth]{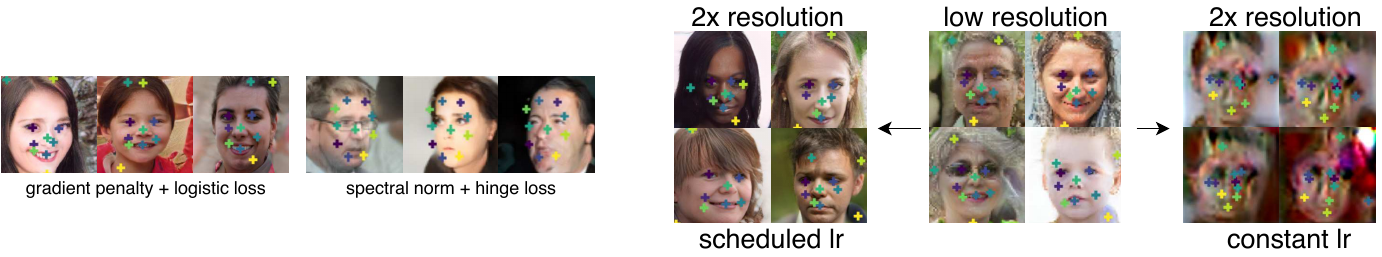}
\end{center}
   \caption{(Left) \textbf{GAN Loss Importance.} Without gradient penality + logistic loss, as in SPADE, keypoint coordinates remain static. (Right) \textbf{Scheduling the keypoint generator learning rate.} Reducing the learning rate after each progressive up-scaling step prevents mode collapse and enables high-resolution training. }
\label{fig:ablation_training}
\end{figure*}

\paragraph{Ablation Test on GAN Loss}
If we replace the generator, discriminator, and gradient peanalty losses introduced in the main paper  with the spectral norm \cite{miyato2018spectral} and hinge loss \cite{miyato2018spectral, park2019semantic} used in the original SPADE architecture, we get mostly static, meaningless  latent keypoint coordinates. The object part location information is entangled with the key part appearance. The comparison is shown in Figure~\ref{fig:ablation_training}.

\paragraph{Ablation Test on Background Loss}
If we remove the background loss, most keypoints are still at reasonable positions while some move to the background. As shown in Figure~\ref{fig:ablation_bg_loss}, the yellow keypoint is on the background while all the others are still on the foreground. 

\paragraph{Removing Keypoint Scheduler.} If we move the keypoint scheduler, i.e., updating keypoint generator during resolution adaption, the keypoint locations diverge and the appearance collapses, as shown in Figure~\ref{fig:ablation_training}.

\section{Failure Cases and Limitations} 
\begin{figure*}[h]
% \vspace{-5pt}
\begin{center}
   \includegraphics[width=0.98\linewidth]{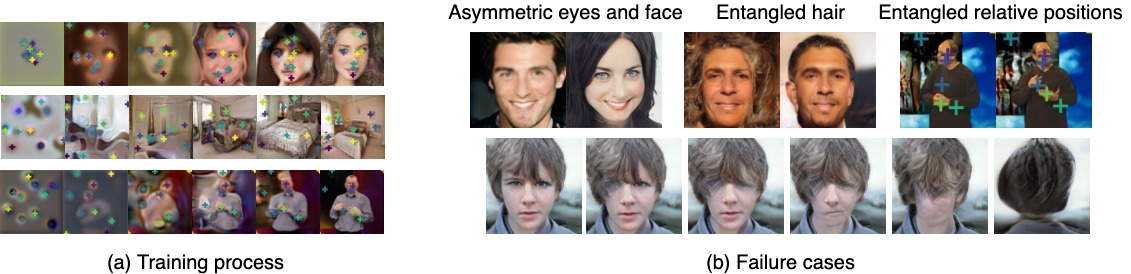}
\end{center}
   \caption{(a) \textbf{Training process.} We visualize the image generated during the training. (b) \textbf{Failure cases.} The left top two images show asymmetric faces: different eye colors for the man and different blusher for the woman. The middle top two images show the entanglement of hair and background. The right top two images show that the pose of head is hidden in the relative positions of other keypoints than the keypoints on the head. We visualize the process of removing parts at the bottom. We sequentially remove the left eye, right eye, mouth, nose, and the entire face. Due to the entanglement of hair and background, the hair remains even if we remove the whole face.}
\label{fig:failure}
% \vspace{-5pt}
\end{figure*}

As described in the main text, our model sometimes generates asymmetric faces as shown in the first two images in Figure~\ref{fig:failure}. In addition, the hair sometimes is entangled with the background, especially long hair, as shown in the right two images in Figure~\ref{fig:failure}. 

From the ablation tests in training strategy, we can see that this method can heavily depend on the state-of-the-art GAN training loss function and image-to-image translation architectures. In fact, we observed some image quality degeneration as the training goes on in the highest resolution ($512\times 512$). Therefore, we apply early stopping in the highest resolution. We expect that researchers will push GAN and spatially adaptive image-to-image translation even further. We believe that our LatentKeypointGAN would directly benefit from these advances.
%In this paper, we only provide the idea and focus on unsupervised local image editing without using any loss on pairs of images.

\section{Ethics Statement}

This research provides a new unsupervised disentanglement and editing method. By contrast to existing supervised ones, e.g., those requiring manually annotated masks, ours can be trained on any image collection. This enables training on very diverse sets as well as on personalized models for a particular population and can thereby counter biases in the annotated datasets. 

Photo-realistic editing tools have the risk of abuse via \emph{deep fakes}. A picture of a real person could be altered to express something not intended by the person. In that regard, our method has the advantage of only enabling the editing of generated images; it does not enable modifying real images; it only works in the synthetic to real direction. However, future work could extend it with an appearance encoder, which bears some risk.

Another risk is that the network could memorize appearances from the training set and therefore re-produce unwanted deformations of the pictured subjects.
While \cite{brock2018large} and \cite{karras2018progressive} argue that GANs do not memorize training datasets, recently \cite{Feng_2021_ICCV} empirically proved that whether GANs memorize training datasets depends on the complexity of the training datasets. Therefore, our model has some risk of leaking such identity information if the training set is very small or the number of persons involved is limited, such as BBCPose.
% Due to the nature of GANs, where the discriminator and generator learn how faces look but are not designed to memorize entire faces, our method has a low risk of leaking such identity information unless the training set is very small. However, we noted that for the BBCpose dataset, which only contains a handful of subjects, the source appearance is learned.
%Still, it is unclear whether an adversarial could receive more information. 
To mitigate the risk, we only use established and publicly available datasets, in particular those that collect pictures of celebrities or other public figures and also those not containing any person (bedroom dataset).

Our IRB approved the survey and we collected explicit and informed consent at the beginning of the study. Results are stored on secure servers. This research did not cause any harm to any subject. 

%On the technical side, we disclose all limitations in both the main paper and the supplementary. This research does not have conflicts with others. 
%We credit the previous works in the Introduction (Section \ref{sec:intro}) and Related Work (Section \ref{sec:related_work}). 
%This research does not harm anyone's privacy. 
%We do not leak any information about the people who did the survey unless they 
%told us that want to be acknowledged publicly. 
%All the datasets used in this research are publicly available. No confidential information is used throughout the whole research. Our keypoint detector is trained on celebrities. We do not train any kind of encoders on ordinary people. Our method can only generate fake face, such that no facial identity will be leaked. A possible risk is that people train an encoder for our model to edit existing people's face.

%\section{Reproducibility}
%We not only described our algorithm in Section~\ref{sec:method} but also provide detailed information of our architecture, training strategy, and hyperparameters in the supplementary~\ref{supp:archi_details}. All the datasets we used are publicly available and the pre-processing is also described in Section~\ref{experiments}. Furthermore, we provide the source code to facilitate future work in this direction of GAN-based generative image editing.

% \bibliographystyle{IEEEtran}
% \bibliography{sn-bibliography}

% that's all folks
\end{document}